\documentclass[lettersize,journal]{IEEEtran}
\usepackage{amsmath,amsfonts}
\usepackage{algorithmic}
\usepackage{algorithm}
\usepackage{array}
\usepackage[caption=false,font=normalsize,labelfont=sf,textfont=sf]{subfig}
\usepackage{textcomp}
\usepackage{stfloats}
\usepackage{url}
\usepackage{verbatim}
\usepackage{graphicx}
\usepackage{cite}
\newcommand{\proposedmethod}{MIRROR}
\usepackage{booktabs}
\usepackage{multirow}
\usepackage[hidelinks]{hyperref}
\usepackage{makecell}

\usepackage{amsthm}
\theoremstyle{plain}
\newtheorem{proposition}{Proposition}

\begin{document}

\bstctlcite{IEEEexample:BSTcontrol}

\title{\proposedmethod: Multi-Modal Pathological Self-Supervised Representation Learning via Modality Alignment and Retention}

\author{Tianyi Wang, Jianan Fan, Dingxin Zhang, Dongnan Liu, Yong Xia, Heng Huang, and Weidong Cai
\thanks{This work was supported in part by China Scholarship Council (CSC) -- The University of Sydney (USYD) Postgraduate Research Scholarship.}
\thanks{T. Wang, J. Fan, D. Zhang, D. Liu and W. Cai are with the School of Computer Science, The University of Sydney, Sydney, NSW, 2006, Australia (e-mail: {twan0134, jfan6480, dzha2344}@uni.sydney.edu.au, {dongnan.liu, tom.cai}@sydney.edu.au).}
\thanks{Y. Xia is with the National Engineering Laboratory for Integrated Aero-Space-Ground-Ocean Big Data Application Technology, School of Computer Science and Engineering, Northwestern Polytechnical University, Xi’an, 710072, China, with Research \& Development Institute of Northwestern Polytechnical University in Shenzhen, Shenzhen 518057, China, and also with the Ningbo Institute of Northwestern Polytechnical University, Ningbo 315048, China (e-mail: yxia@nwpu.edu.cn).}
\thanks{H. Huang is with the Department of Computer Science, University of Maryland, College Park, MD 20742, USA (e-mail: heng@umd.edu).}
\thanks{W. Cai is the corresponding author.}}

\maketitle

\begin{abstract}
Histopathology and transcriptomics are fundamental modalities in cancer diagnostics, encapsulating the morphological and molecular characteristics of the disease. Multi-modal self-supervised learning has demonstrated remarkable potential in learning pathological representations by integrating diverse data sources. Conventional multi-modal integration methods primarily emphasize modality alignment, while paying insufficient attention to retaining the modality-specific intrinsic structures. However, unlike conventional scenarios where multi-modal inputs often share highly overlapping features, histopathology and transcriptomics exhibit pronounced heterogeneity, offering orthogonal yet complementary insights. Histopathology data provides morphological and spatial context, elucidating tissue architecture and cellular topology, whereas transcriptomics data delineates molecular signatures through quantifying gene expression patterns. This inherent disparity introduces a major challenge in aligning these modalities while maintaining modality-specific fidelity. To address these challenges, we present \proposedmethod, a novel multi-modal representation learning framework designed to foster both modality alignment and retention. \proposedmethod\space employs dedicated encoders to extract comprehensive feature representations for each modality, which is further complemented by a modality alignment module to achieve seamless integration between phenotype patterns and molecular profiles. Furthermore, a modality retention module safeguards unique attributes from each modality, while a style clustering module mitigates redundancy and enhances disease-relevant information by modeling and aligning consistent pathological signatures within a clustering space. Extensive evaluations on The Cancer Genome Atlas (TCGA) cohorts for cancer subtyping and survival analysis highlight \proposedmethod's superior performance, demonstrating its effectiveness in constructing comprehensive oncological feature representations and benefiting the cancer diagnosis. Code is available at \url{https://github.com/TianyiFranklinWang/MIRROR}.
\end{abstract}


\section{Introduction}
\label{sec:introduction}

\IEEEPARstart{H}{istopathology} images are widely regarded as the gold standard in cancer diagnosis, offering critical insights into the presence, type, grade, and prognosis of cancer~\cite{chang2007evidence}. These images encapsulate a wealth of morphological features that serve as the foundation of cancer diagnostics~\cite{7886294,chen2022scaling,chen2024towards,xiang2022dsnet,li2023single}. Meanwhile, advancements in high-throughput sequencing technologies, such as polymerase chain reaction (PCR)~\cite{saiki1985enzymatic}, have further expanded oncological diagnostic capabilities by enabling the analysis of molecular data, including transcriptomics data, which delineates gene expression profiles, offering molecular signatures of the disease. The integration of the morphological and molecular modalities within multi-modal diagnostics significantly enhances the accuracy of cancer diagnosis and prognosis prediction, providing a more comprehensive and precise understanding of the disease.

\begin{figure}[!t]
\centering
\includegraphics[width=\columnwidth]{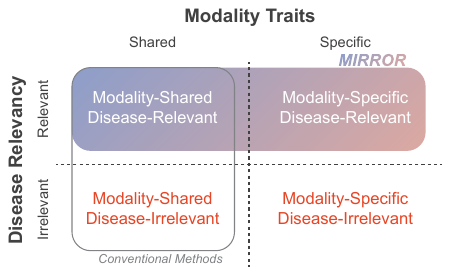}
\caption{\textbf{\proposedmethod\space compared with conventional multi-modal integration methods.} Unlike conventional methods that primarily emphasize capturing modality-shared information while paying limited attention to modality-specific intrinsic structures and indiscriminately learning both disease-relevant and irrelevant data with high redundancy, \proposedmethod\space is specifically designed to balance modality alignment and retention. By selectively preserving only disease-relevant features, it effectively mitigates redundancy, thereby enhancing the model’s efficiency and representational capability.}
\label{fig:intro}
\end{figure}

Despite the transformative potential of multi-modal diagnostics, several challenges hinder their widespread adoption. The incorporation of molecular data into existing pathology workflows presents substantial complexities, further exacerbating the workload for already overburdened pathologists. Additionally, the scarcity of annotated paired data, due to the resource-intensive and time-consuming nature of labeling, significantly hinders the adoption of supervised learning methods. In this context, multi-modal self-supervised learning (SSL) emerges as a promising alternative, offering the capability to capture robust and comprehensive oncological feature representations without reliance on extensive annotations. Multi-modal SSL methods~\cite{radford2021learning,wang2022medclip,xu2024multimodal} have demonstrated remarkable success in both natural and medical domains, effectively aligning modalities such as image-text pairs. However, the direct extension of such aligning techniques to histopathology and transcriptomics data presents distinct challenges. Unlike conventional use cases, where multi-modal inputs often exhibit highly overlapping features, histopathology and transcriptomics data pairs are inherently more heterogeneous, as they operate at different biological scales and encode distinct yet complementary dimensions of disease-related information. Histopathology provides a morphological and spatial view of tissue architecture, capturing phenotypic traits, while transcriptomics quantifies gene expression levels, uncovering the molecular processes and pathways underlying the disease. Although there are shared correlations between these modalities, each modality also retains substantial modality-specific information~\cite{chang2007evidence,boehm2022harnessing}. As shown in Figure~\ref{fig:intro}, existing multi-modal methods~\cite{jaume2024transcriptomics,zhang2024prototypical,jaume2024modeling,vaidya2025molecular} primarily focus on aligning shared information between modalities, while giving comparatively less attention to the rich and modality-specific information inherent to each data type. For instance, in~\cite{vaidya2025molecular}, the authors employed an encoder-only architecture with contrastive learning to enforce representation alignment. However, the optimum training target will be achieved when representations from the same sample become indistinguishable across modalities, thereby eliminating essential modality-specific attributes. Moreover, both histopathology and transcriptomics data contain redundant, disease-unrelated information, including repetitive structural patterns and genes with overlapping functions or pathways. Reducing such redundancies can enable the model to extract more clinically meaningful, disease-relevant representations. Beyond these challenges, the inherent heterogeneity in data formats presents an additional layer of complexity. The histopathology data is structured as 2D image patches, whereas transcriptomic data is represented as tabular numerical values, necessitating careful architectural design.

To address these challenges, we propose \proposedmethod\space (\textbf{M}ulti-modal patholog\textbf{I}cal self-supe\textbf{R}vised \textbf{R}epresentation learning via m\textbf{O}dality alignment and \textbf{R}etention), a novel multi-modal SSL framework designed to foster both modality alignment and retention. \proposedmethod\space adopts dedicated Transformer-based~\cite{vaswani2017attention} encoders to extract rich and discriminative feature representations for each modality while being specifically tailored to accommodate the inherent heterogeneity in data format. To facilitate seamless integration within the latent space, a modality alignment module is introduced, dynamically drawing paired data into closer proximity while dispersing unrelated samples. To safeguard modality-specific fidelity, a modality retention module is employed to ensure the preservation of unique modality attributes. This module challenges the model to maintain modality-specific intrinsic structures by reconstructing key features after perturbation. Additionally, to mitigate redundancy and enhance disease-relevant information, \proposedmethod\space incorporates a style clustering module, which maps feature embeddings into a statistical space to capture consistent pathological styles while minimizing intra-modality redundancy. Subsequently, a prototype clustering mechanism further aligns the captured styles in the clustering space, mitigating inter-modality redundancy and reinforcing biologically meaningful correspondences. Together, these synergistic modules cultivate a well-structured representation space by disentangling and preserving both modality-shared and modality-specific signatures while suppressing irrelevant variations, enabling \proposedmethod\space to deliver robust and comprehensive multi-modal representations, advancing the capabilities of multi-modal diagnostics.

Furthermore, the vast number of genes available in transcriptomics data presents a significant challenge in identifying those most relevant to disease development. \proposedmethod\space addresses this issue through a novel preprocessing pipeline that integrates both machine learning-driven feature selection with biological knowledge to distill high-dimensional transcriptomics data, creating refined and disease-focused transcriptomics datasets.

The proposed method is evaluated using 5-fold cross-validation on multiple cohorts from the TCGA dataset~\cite{tomczak2015review}, targeting critical downstream tasks including cancer subtyping and survival analysis. The evaluation incorporates both linear probing and few-shot learning settings to comprehensively assess the performance and generalizability of the model. The key contributions of this study are outlined as follows:
\begin{itemize}
    \item \proposedmethod, a novel multi-modal SSL model, is designed to facilitate both modality alignment and retention, enabling the effective preservation of both modality-shared and modality-specific information.
    \item A consistent pathological style-based clustering mechanism is introduced to preserve disease-relevant information while mitigating redundancy.
    \item A novel preprocessing pipeline for transcriptomics data is proposed, integrating machine learning-driven feature selection with biological knowledge to create refined transcriptomics datasets.
    \item Comprehensive evaluations are conducted across diverse cohorts from the TCGA dataset, focusing on cancer subtyping and survival analysis tasks, substantiating the superior performance and effectiveness of the proposed approach.
\end{itemize}

\section{Related Work}

\subsection{Self-Supervised Learning in Computer Vision}
Recent advancements in SSL in computer vision (CV) have significantly reduced the reliance on labeled data while achieving performance comparable to, or even surpassing, that of supervised methods. SSL can be categorized into three main styles: contrastive learning~\cite{chen2020simple,he2020momentum}, generative learning~\cite{bao2021beit,he2022masked}, and hybrid methods~\cite{zhou2021ibot,oquab2023dinov2} that integrate both approaches. Specifically, SimCLR~\cite{chen2020simple} achieves contrastive learning by maximizing agreement between different augmented views of the same image, while MoCo~\cite{he2020momentum} introduces a momentum encoder and a dynamic memory queue to improve stability. BEiT~\cite{bao2021beit} leverages discrete variational autoencoders to tokenize images and reconstruct visual tokens, whereas MAE~\cite{he2022masked} directly reconstructs raw pixels from masked patches. Hybrid methods such as iBOT~\cite{zhou2021ibot} combine contrastive and reconstruction on both \textit{[CLS]} and patch tokens, and DINOv2~\cite{oquab2023dinov2} incorporates SwAV-style~\cite{caron2020unsupervised} prototype learning with scalable training strategies.

Multi-modal SSL has further enhanced the capability of models to integrate and align data from different modalities. Among various inputs, vision-language learning~\cite{radford2021learning,jia2021scaling,alayrac2022flamingo} is the most extensively studied and demonstrates exceptional performance. Due to the largely overlapping information between image and text data, aligning the two modalities in the latent space alone is often sufficient to achieve impressive results. CLIP~\cite{radford2021learning} jointly trains an image encoder and a text encoder to maximize the similarity between paired image-text samples, while minimizing it for unpaired ones. ALIGN~\cite{jia2021scaling} extends this training strategy by incorporating noisy image-text pairs, demonstrating the robustness of contrastive alignment at scale. Flamingo~\cite{alayrac2022flamingo} builds on this by introducing cross-attention modules that allow the model to condition visual representations directly on textual context, enabling effective few-shot learning across modalities.

\subsection{Pathological Self-Supervised Learning}
To overcome the scarcity of labeled data in Computational Pathology (CPath), a large amount of prior works~\cite{liu2020pdam,lin2022label,li2023task,fan2024revisiting,fan2024seeing,fan2025structuring} have explored unsupervised learning and transfer learning techniques. In recent years, SSL methods have gained increasing recognition for their superior potential in CPath, particularly in leveraging large-scale unlabeled Whole Slide Images (WSIs). Similar to general CV, SSL approaches in CPath can be broadly classified into three categories: contrastive~\cite{chen2022scaling,wang2022transformer,ciga2022self,ahmed2024pathalign,sun2024cpath}, generative~\cite{yang2023mrm,lu2023multi}, and hybrid~\cite{chen2024towards,xu2024whole,vorontsov2023virchow,koohbanani2021self}. In the contrastive learning paradigm, HIPT~\cite{chen2022scaling} introduces a hierarchical contrastive strategy tailored to multi-resolution WSIs, while CTransPath~\cite{wang2022transformer} combines CNN and Transformer backbones to enhance feature generalizability. PathAlign~\cite{ahmed2024pathalign} and CPath-Omni~\cite{sun2024cpath} extend contrastive SSL to multi-modal contexts by aligning histopathology images with textual descriptions or clinical reports. Among generative methods, MRM~\cite{yang2023mrm} proposes masked relation modeling for joint pretraining on medical images and genomics, whereas MMP-MAE~\cite{lu2023multi} leverages mixed attention and modality-specific masked modeling to integrate differently stained images. Hybrid approaches seek to combine the strengths of both paradigms. UNI~\cite{chen2024towards}, GigaPath~\cite{xu2024whole}, and Virchow~\cite{vorontsov2023virchow} adopt DINOv2-based~\cite{oquab2023dinov2} self-distillation frameworks and scale up to millions of pathology slides, aiming to establish general-purpose foundation models. Self-Path~\cite{koohbanani2021self} employs multi-task self-supervised learning to improve representation quality in low-label regimes.

Multi-modal pathological SSL enhances the model's ability to learn comprehensive pathological representations by integrating diverse inputs from the diagnostic process. This includes combining WSIs with clinical reports or text captions~\cite{singhal2023large,lu2024visual,li2024llava}, incorporating transcriptomics data~\cite{jaume2024transcriptomics}, and fusing images from different staining techniques~\cite{lu2023multi,jaume2024multistain}. CONCH~\cite{lu2024visual} leverages the CoCa~\cite{yu2022coca} framework to train on paired WSI-text data, while LLaVA-Med~\cite{li2024llava} adapts a vision-language assistant for biomedical instruction tuning. MADELEINE~\cite{jaume2024multistain} improves the model’s ability to interpret differently stained images by jointly pretraining on Hematoxylin and Eosin (H\&E) and Immunohistochemistry (IHC) slides. However, these methods often overlook the inherent redundancy in WSIs and other inputs, leading to the indiscriminate learning of both disease-relevant and disease-irrelevant information, which hampers the models' performance and efficiency.

\subsection{Histopathology and Transcriptomics Multi-Modal Learning}
Multi-modal integration methods~\cite{zhang2024prototypical,guo2024histgen,chen2020pathomic,jaume2024transcriptomics,zheng2024graph,vaidya2025molecular,wang2025histo} that combine histopathology and transcriptomics have shown impressive capabilities in various pathological tasks, including cancer diagnosis and survival analysis. 

In practice, these integration methods predominantly emphasize bridging these modalities through contrastive learning or alternative alignment techniques, such as clustering and cross-attention, to merge data from multiple sources into a unified feature space. TANGLE~\cite{jaume2024transcriptomics} proposed an SSL framework that aligns histopathology slides and transcriptomic profiles through contrastive learning, opening new avenues for multi-modal representation learning in pathology. THREADS~\cite{vaidya2025molecular} extended the scope of data modalities to include DNA, alongside RNA and histology, in a foundation model that still primarily relies on contrastive learning. PIBD~\cite{zhang2024prototypical} employed prototypical clustering and disentanglement strategies to align histology and pathway-level transcriptomics representations. Pathomic Fusion~\cite{chen2020pathomic} proposed a fusion framework leveraging gated attention and matrix multiplication to integrate WSIs, spatial cell graphs, and genomic features. Zheng \textit{et al.}~\cite{zheng2024graph} applied graph attention networks to combine image and gene expression features, and G-HANet~\cite{wang2025histo} introduced a cross-attention mechanism to dynamically associate histological and genomic representations.

While recent advancements have yielded improved outcomes, it is important to recognize that these modalities, despite sharing some information, also contain a vast amount of comprehensive and distinct insights into the disease, reflecting their inherent heterogeneity. However, current methods often overlook the rich and diverse signals embedded in the unique characteristics of each modality. Xing \textit{et al.}~\cite{xing2024comprehensive} made a notable advancement by introducing a saliency-aware masking strategy with adaptive distillation, effectively preserving discriminative, modality-specific knowledge and demonstrating the strong potential of masked modeling for nuanced multi-modal representation learning.

\begin{figure*}[!t]
\centering
\includegraphics[width=\textwidth]{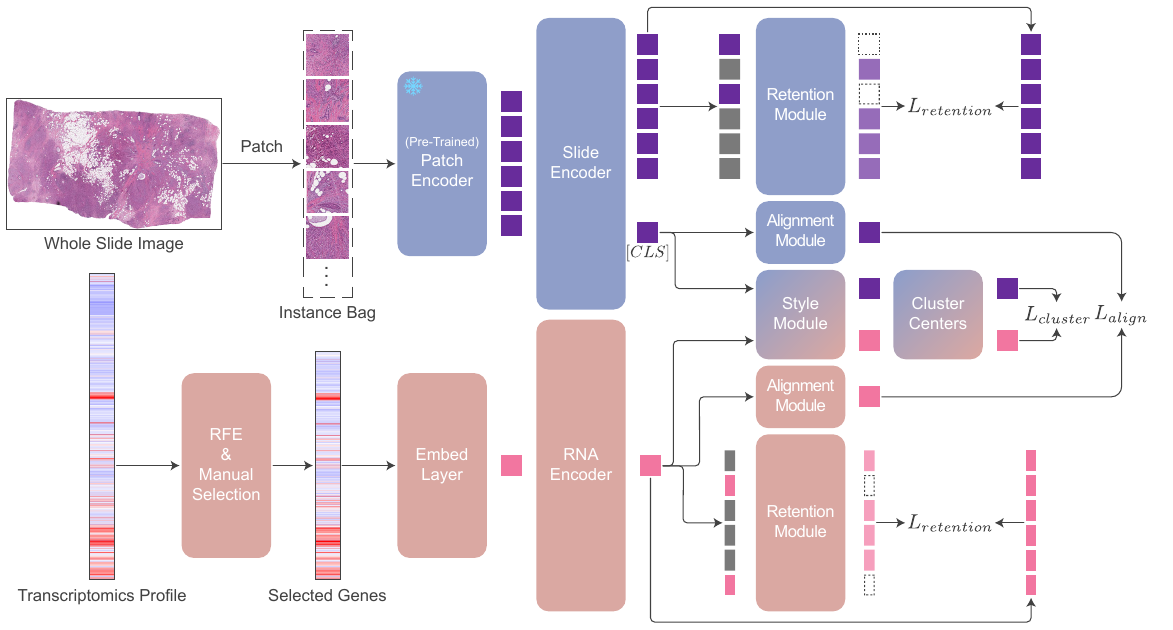}
\caption{\textbf{Overview of \proposedmethod.} WSIs are first partitioned into patches, which are processed through a pre-trained patch encoder to extract patch-level feature representations. These features are subsequently aggregated by the slide encoder to encapsulate slide-level characteristics into a \textit{[CLS]} token while projecting patch embeddings into the shared pathological latent space. Transcriptomics data are preprocessed using RFE and manual selection to identify high disease-related genes. The refined transcriptomic features are then embedded into a compact representation and mapped into the shared latent space via an RNA encoder. An alignment module for each modality aligns representations across modalities, guided by the alignment loss ($L_{align}$). Meanwhile, modality-specific retention modules utilize perturbed inputs from both encoded patch and transcriptomics features to capture modality-specific intrinsic structures, contributing to the retention loss ($L_{retention}$). Finally, both slide and transcriptomics representations are processed through a style clustering module to learn and compare their pathological styles against learnable cluster centers, with the clustering loss ($L_{cluster}$) used to align consistent pathological styles within the cluster space.}
\label{fig:method_diagram}
\end{figure*}

\section{Methodology}
As illustrated in Figure~\ref{fig:method_diagram}, \proposedmethod\space consists of four main components, each described in detail below.

\subsection{Modality Preprocessing and Encoding}
\proposedmethod~first applies tailored preprocessing strategies to histopathology and transcriptomics data, including a novel gene selection pipeline for transcriptomics. The processed inputs are then encoded using Transformer-based models to project them into the shared pathological latent space effectively.

\subsubsection{Slide Encoder}
WSIs are first partitioned into patches, forming an instance bag that is processed through a pre-trained patch encoder to extract patch-level feature representations, denoted as $\mathbf{P} \in \mathbb{R}^{N \times D_p}$, where $N$ is the number of patches in the instance bag and $D_p$ is the dimensionality of the patch-level representations. These representations are subsequently fed into the slide encoder $f$ to obtain slide-level representations of patch tokens and a global slide \textit{[CLS]} token:
\begin{equation} 
\begin{split}
    \mathbf{S}, \mathbf{S}^{\textit{[CLS]}} = f(\mathbf{P}),
\end{split} 
\end{equation}
where $\mathbf{S} \in \mathbb{R}^{N \times D}$ represents the slide-level feature embeddings for the patch tokens, $D$ is the dimensionality of the shared latent space, and $\mathbf{S}^{\textit{[CLS]}}$ is the class token capturing global slide-level information. To achieve effective positional encoding and attention, our approach incorporates a modified version of TransMIL~\cite{shao2021transmil}, which includes two attention blocks and a Pyramid Position Encoding Generator (PPEG) module.

\subsubsection{Transcriptomics Preprocessing}
\begin{figure}[!t]
\centering
\includegraphics[width=\columnwidth]{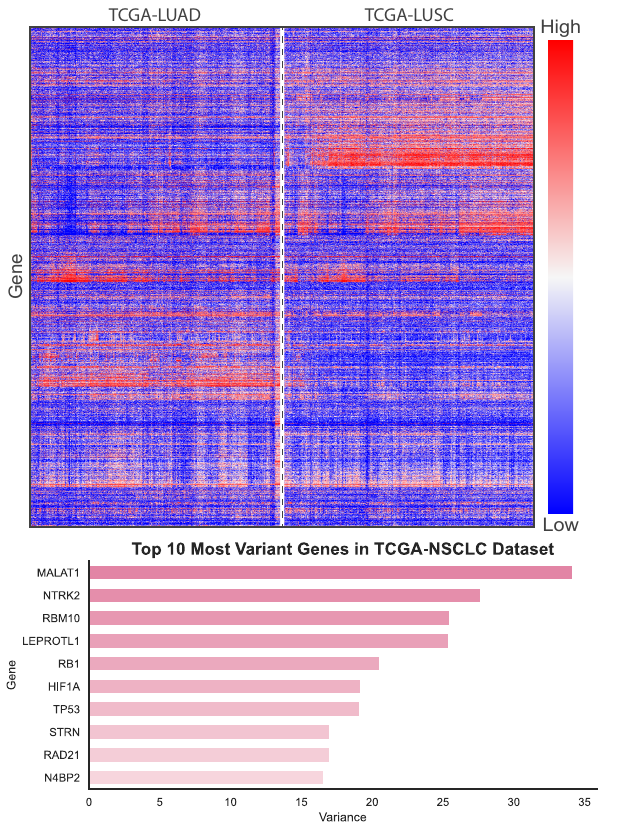}
\caption{\textbf{Transcriptomics data distributions in the TCGA-NSCLC dataset.} The top panel displays a heatmap visualization of transcriptomic data for two subtypes in TCGA-NSCLC: TCGA-LUAD on the left and TCGA-LUSC on the right. The data distribution exhibits substantial variability and clear subtype distinction after preprocessing, providing a robust foundation for representation learning. The bottom bar plot highlights the top 10 most variant genes in the TCGA-NSCLC dataset, identified with reference to the COSMIC database, demonstrating extraordinary biological explainability.}
\label{fig:transcriptomics_heatmap}
\end{figure}

Transcriptomics profiles are inherently high-dimensional, encompassing a vast number of genes, many of which exhibit redundancy or limited relevance to oncogenic processes. Thus, effective gene selection is crucial for achieving optimal performance. \proposedmethod\space employs a hybrid gene selection strategy that integrates both machine learning-driven feature selection with biologically curated gene filtering to address this challenge. To identify the most discriminative genes, recursive feature elimination (RFE)~\cite{guyon2002gene} is utilized to determine a highly performant support set. RFE iteratively trains a predictive model $\psi: \mathbb{R}^{D_g} \rightarrow \mathbb{R}$ on the raw transcriptomics input matrix $\mathbf{T}_{\text{raw}} \in \mathbb{R}^{N_s \times D_g}$, where $N_s$ denotes the number of samples and $D_g$ is the total number of genes. The importance of each gene is quantified by the squared magnitude of the model’s learned coefficients $\boldsymbol{\beta} = [\beta_1, \beta_2, \dots, \beta_{D_g}]^\top$:
\begin{equation}
    I_g = |\beta_g|^2.
\end{equation}
At each iteration, the gene with the lowest importance score is eliminated:
\begin{equation}
g^* = \arg\min_{g} I_g,
\end{equation}
and the model is retrained on the reduced feature set until only $K$ features remain, yielding a compact subset of highly informative genes.
Additionally, to further ensure interpretability and biological relevance, manually curated genes associated with specific cancer subtypes are selected based on the COSMIC database~\cite{sondka2024cosmic}. This dual strategy ensures a balance between model performance and biological relevance as shown in Figure~\ref{fig:transcriptomics_heatmap}.

\subsubsection{RNA Encoder}
The selected genes, although significantly reduced, remain numerous for direct encoding. To address this, these genes are passed through an embedding layer to reduce their dimensionality, producing a compact representation denoted as $\hat{\mathbf{T}} \in \mathbb{R}^{D_t}$, where $D_t$ is the dimensionality of the compact representation. To model intricate gene-gene interactions and extract biologically meaningful transcriptomic representations, \proposedmethod\space utilizes a Transformer-based RNA encoder~\cite{vaswani2017attention}. This encoder incorporates a learnable gene encoding token $\mathbf{G} \in \mathbb{R}^{D_t}$, which encapsulates the inherent correlations among gene expression patterns. The final encoded transcriptomic representation is computed as:
\begin{equation} 
\mathbf{T} = g(\hat{\mathbf{T}}, \mathbf{G}),
\end{equation}
where $\mathbf{T} \in \mathbb{R}^{D}$ represents the encoded transcriptomics features in the shared latent space, and $D$ is the dimensionality, matching the output dimensionality of the slide encoder.

\subsection{Preliminaries}
\label{subsec:preliminaries}
As discussed in Section~\ref{sec:introduction}, the output of each encoder can be conceptually 
decomposed into four parts: 
(1)~disease-relevant, modality-shared; 
(2)~disease-relevant, modality-specific; 
(3)~disease-irrelevant, modality-shared; and 
(4)~disease-irrelevant, modality-specific. 
Hence, for the $i$-th paired sample, the outputs of the slide and RNA encoders can be written as:
\begin{equation}
\begin{split}
    \mathbf{S}^i &= S_{\mathrm{r},s}^i + S_{\mathrm{r},u}^i + S_{\mathrm{i},s}^i + S_{\mathrm{i},u}^i, \\
    \mathbf{T}^i &= T_{\mathrm{r},s}^i + T_{\mathrm{r},u}^i + T_{\mathrm{i},s}^i + T_{\mathrm{i},u}^i,
\end{split}
\end{equation}
where ``$\mathrm{r}$'' (relevant) vs.\ ``$\mathrm{i}$'' (irrelevant) indicates disease relevance, and ``$s$'' (shared) vs.\ ``$u$'' (unique)  indicates modality traits. In this notation, the ``$+$'' denotes the composition of subspace features rather than simple vector addition. While we speak of disease relevance for clarity, in practice, these relevant subspaces can capture any shared pathological characteristics that drive meaningful variation 
in the data. Additionally, for notational conciseness, $\mathbf{S}$ and $\mathbf{S}^{\textit{[CLS]}}$ are treated equivalently in this context.

\subsection{Modality Alignment Module}
The modality alignment module aims to map each encoder output to a common latent space where modality-shared components are systematically brought closer for paired samples, while pushing away irrelevant or mismatched components. To extract the modality-shared information, the encoded representations $\mathbf{S}^{\textit{[CLS]}}$ and $\mathbf{T}$ are mapped into a shared latent space using a modality alignment module for each modality, defined as:
\begin{equation}
\begin{split}
    \mathbf{S_{\text{align}}} & = f_{\text{align}}(\mathbf{S}^{\textit{[CLS]}}), \\
    \mathbf{T_{\text{align}}} & = g_{\text{align}}(\mathbf{T}),
\end{split}
\end{equation}
where $\mathbf{S_{\text{align}}}, \mathbf{T_{\text{align}}} \in \mathbb{R}^{D}$ are the aligned representations containing modality-shared information defined as:
\begin{equation}
\begin{split}
    \mathbf{S}_{\text{align}}^i & = S_{\mathrm{r},s}^i + S_{\mathrm{i},s}^i, \\
    \mathbf{T}_{\text{align}}^i & = T_{\mathrm{r},s}^i + T_{\mathrm{i},s}^i.
\end{split}
\end{equation}

To guide the alignment process, an alignment loss function is employed. Inspired by \cite{oord2018representation}, the loss function is formulated as:
\begin{equation}
\begin{split}
    L_{\text{align}} & = -\frac{1}{2B} \sum_{i=1}^{B} \log \left( 
        \frac{\exp(\tau {\mathbf{S}_{\text{align}}^{i}}^\top \mathbf{T}_{\text{align}}^{i})}
             {\sum_{j=1}^{B} \exp(\tau {\mathbf{S}_{\text{align}}^{i}}^\top \mathbf{T}_{\text{align}}^{j})}
    \right) \\
    & \quad -\frac{1}{2B} \sum_{i=1}^{B} \log \left( 
        \frac{\exp(\tau {\mathbf{T}_{\text{align}}^{i}}^\top \mathbf{S}_{\text{align}}^{i})}
             {\sum_{j=1}^{B} \exp(\tau {\mathbf{T}_{\text{align}}^{i}}^\top \mathbf{S}_{\text{align}}^{j})}
    \right),
\end{split}
\end{equation}
where $B$ is the batch size, and $\tau$ is a temperature parameter implemented as a learnable scalar, jointly optimized with the model parameters.

This loss function inherently encourages high similarity between paired samples $(\mathbf{S}_{\text{align}}^{i}, \mathbf{T}_{\text{align}}^{i})$ while reducing the similarity between negative pairs $(\mathbf{S}_{\text{align}}^{i}, \mathbf{T}_{\text{align}}^{j})$ for $i \neq j$. At convergence, the following conditions hold:
\begin{equation}
    \mathbf{S}_{\text{align}}^{i} \approx \mathbf{T}_{\text{align}}^{i} \quad \text{and} \quad \mathbf{S}_{\text{align}}^{i} \cdot \mathbf{T}_{\text{align}}^{j} \rightarrow 0 \quad \forall i \neq j,
\end{equation}
where the dot product measures similarity, ensuring that only corresponding pairs exhibit high alignment while unrelated pairs are nearly orthogonal in the shared latent space.

Through this iterative optimization process, the model fosters the formation of distinct clusters, wherein samples sharing similar pathological signatures naturally coalesce in the latent space. Concurrently, the enforced separation between dissimilar instances ensures the emergence of well-differentiated clusters, thereby enhancing the model's capacity to capture discriminative features.

\subsection{Modality Retention Module}
In contrast to the conventional SSL scenarios where input modalities often exhibit highly overlapping or shared features, histopathology and transcriptomics data are fundamentally heterogeneous. This heterogeneity means that simply aligning these two modalities risks discarding essential modality-specific information encoded in $S_{\mathrm{r},u}^i, S_{\mathrm{i},u}^i, T_{\mathrm{r},u}^i$, and $T_{\mathrm{i},u}^i$. To address this, \proposedmethod\space introduces a modality retention module that is explicitly designed to safeguard unique attributes of each modality without compromising shared semantics.

For histopathology, \proposedmethod\space employs a masked patch modeling task. A subset of patch tokens from $\mathbf{S}$ is randomly masked, producing a corrupted representation $\mathbf{S}_{\text{masked}}$, from which the retention module tries to reconstruct the missing tokens by learning the mapping:
\begin{equation}
    \mathbf{S}_{\text{retention}} = f_{\text{retention}}(\mathbf{S}_{\text{masked}}).
\end{equation}
The reconstructed representation therefore encapsulates modality-specific semantic information while preserving the integrity of shared components:
\begin{equation}
    \mathbf{S}_{\text{retention}} \approx \mathbf{S} = S_{\mathrm{r},s} + S_{\mathrm{r},u} + S_{\mathrm{i},s} + S_{\mathrm{i},u}.
\end{equation}

For transcriptomics data, a novel masked transcriptomics modeling task is introduced. A subset of gene representations from $\mathbf{T}$ is randomly masked, forming $\mathbf{T}_{\text{masked}}$. The transcriptomics retention module then reconstructs the masked representations by learning:
\begin{equation}
    \mathbf{T}_{\text{retention}} = g_{\text{retention}}(\mathbf{T}_{\text{masked}}).
\end{equation}
Similarly, the retention incorporates both modality-specific and shared information:
\begin{equation}
    \mathbf{T}_{\text{retention}} \approx \mathbf{T} = T_{\mathrm{r},s} + T_{\mathrm{r},u} + T_{\mathrm{i},s} + T_{\mathrm{i},u}.
\end{equation}

To ensure the retention of modality-specific information, the representations produced by each encoder should encapsulate rich and holistic semantic information. Consequently, the loss function for modality retention is defined as:
\begin{equation}
\begin{split}
    L_{\text{retention}} & = \frac{1}{2B} \sum_{i=1}^{B} \text{sim}(\mathbf{S}^{i}, \mathbf{S}_{\text{retention}}^{i}) \\
    & + \frac{1}{2B} \sum_{i=1}^{B} \text{sim}(\mathbf{T}^{i}, \mathbf{T}_{\text{retention}}^{i}),
\end{split}
\label{eq:l_retention}
\end{equation}
where $\text{sim}(\cdot)$ denotes the similarity measurement function, for which Mean Squared Error (MSE) is used. Given the stochastic nature of masking, where any patch or gene can be masked at random, the encoder is compelled to distribute modality-specific intrinsic structures uniformly across the entire representation. This strategy ensures each modality's unique biological and morphological characteristics remain well-preserved, improving overall fidelity of the learned representations. Because $L_{\text{align}}$ takes effect only on the modality-shared components, the preservation of modality-specific components is guided exclusively by $L_{\text{retention}}$.

\subsection{Style Clustering Module}
\label{subsec:style_clustering}

To mitigate redundancy and enhance the representation of disease-relevant information, \proposedmethod~incorporates a style clustering module designed to project both $\mathbf{S}^{\textit{[CLS]}}$ and $\mathbf{T}$ into a shared latent space. In this module, each input is first passed through a multilayer perceptron (MLP) to generate a Gaussian distribution in a compressed latent space, parameterized by a mean vector $\mu \in \mathbb{R}^d$ and a log standard deviation vector $\log \sigma \in \mathbb{R}^d$. A latent code $h \sim \mathcal{N}(\mu, \sigma^2 I)$ is then sampled using the reparameterization trick~\cite{kingma2013auto}. The latent code $h$ is further transformed by a linear projection to produce the final representation $z \in \mathbb{R}^D$, where $D$ is the original input dimension. We denote the resulting representations for histopathology and transcriptomics data as $z_{S}$ and $z_{T}$, respectively, and use them for clustering.

To suppress irrelevant or redundant variability, we regularize the latent distributions to remain close to the standard normal prior $\mathcal{N}(0, I)$ by minimizing the Kullback-Leibler (KL) divergence~\cite{kullback1951information}:
\begin{equation}
\label{eq:L_style}
\begin{split}
    L_{\text{style}} 
    &=\; \mathrm{KL}\bigl(\mathcal{N}(\mu_S, \sigma_S^2 I) \,\big\Vert\, \mathcal{N}(0,I)\bigr) \\
        &+\, \mathrm{KL}\bigl(\mathcal{N}(\mu_T, \sigma_T^2 I) \,\big\Vert\, \mathcal{N}(0,I)\bigr),
\end{split}
\end{equation}
where $(\mu_S, \sigma_S)$ and $(\mu_T, \sigma_T)$ are the Gaussian parameters estimated from the histopathology and transcriptomics input, respectively. This standard normal prior is the most compact and symmetric non-informative choice under variational approximations, as it introduces no bias toward any specific direction in the latent space. Using a more informative or structured prior could encode unintended signals and weaken the latent compression effect~\cite{tishby2000information, alemi2016deep}. We prove in Appendix~\ref{app:style-proof} that Eq.~\eqref{eq:L_style} attains a unique global minimum precisely at:
\begin{equation}
\mu_S=\mu_T=0,
\quad
\sigma_S=\sigma_T=1.
\end{equation}
Every bit that drifts away from the prior therefore increases $L_{\text{style}}$, so the network keeps only information that also helps reduce the total loss, effectively pruning disease-irrelevant noise from the latent space.

To facilitate the alignment of disease-relevant features across modalities, \proposedmethod\space introduces a set of learnable cluster centers $\mathbf{P} \in \mathbb{R}^{K \times D}$, shared across both modalities. Each cluster center $\mathbf{p}_{k}$ of $\mathbf{P}$ is $\ell_2$-normalized to prevent degeneracy. Throughout all experiments, we fix the number of prototypes to $K=3{,}000$, following the default setting in~\cite{caron2020unsupervised,oquab2023dinov2}. This prototype-based design promotes alignment by capturing globally consistent, disease-relevant structure rather than enforcing rigid pairwise matching. Unlike hard alignment methods, which can overfit to sample-level noise and lack contextual coherence, the shared cluster centers are learned across the dataset and act as modality-invariant anchors. The use of soft probabilistic assignments further introduces flexibility, allowing each representation to associate with multiple prototypes. This supports uncertainty modeling and avoids over-constraining the latent space, thereby enabling both alignment and retention of clinically meaningful variation.

Given latent representations $z_S$ and $z_T$, we compute their scaled cosine similarities with the cluster centers $\mathbf{P}$ via a matrix multiplication:
\begin{equation}
    [\mathbf{P} z]_k = \langle \mathbf{p}_k, z \rangle = \|z\|_2 \cos(\theta_k),
\end{equation}
where $\theta_k$ denotes the angle between $z$ and $\mathbf{p}_k$. The resulting vector $\mathbf{P} z$ captures the similarity between the representation and each cluster center. These scores are then normalized via softmax to yield probabilistic assignments:
\begin{equation}
    \mathbf{S}_{\text{cluster}} = \text{softmax}(\mathbf{P} z_S), \quad \mathbf{T}_{\text{cluster}} = \text{softmax}(\mathbf{P} z_T).
\end{equation}
These soft assignments represent the affinity of each sample to the $K$ shared cluster centers. Accordingly, $\mathbf{S}_{\text{cluster}}, \mathbf{T}_{\text{cluster}} \in \mathbb{R}^{B \times K}$, where each row is a $K$-dimensional probability vector for one sample in the batch. The alignment between these assignments is encouraged through a bidirectional KL divergence penalty:
\begin{equation}
\label{eq:L_cluster}
\begin{aligned}
L_{\text{cluster}}
   &= \mathrm{KL}\!\bigl(\mathbf{S}_{\text{cluster}}
                         \,\|\, 
                         \mathbf{T}_{\text{cluster}}\bigr)
      + \mathrm{KL}\!\bigl(\mathbf{T}_{\text{cluster}}
                           \,\|\, 
                           \mathbf{S}_{\text{cluster}}\bigr)\\
   &= 2\,\mathrm{JSD}\!\bigl(\mathbf{S}_{\text{cluster}},
                             \mathbf{T}_{\text{cluster}}\bigr).
\end{aligned}
\end{equation}
Here, $\mathrm{JSD}(\cdot,\cdot)$ denotes the Jensen–Shannon divergence between two probability distributions. We have proved in Appendix~\ref{app:cluster-proof} that $L_{\text{cluster}}$ constitutes an explicit upper bound on the negative mutual information between the style representations, i.e.\ $-I(z_S;z_T)\le g(L_{\text{cluster}})$ for a monotone increasing function $g(\cdot)$. Consequently, minimizing $L_{\text{cluster}}$ increases the mutual information shared by the two modalities.

The total style clustering loss is defined as:
\begin{equation}
L_{\text{style clustering}} = L_{\text{style}} + L_{\text{cluster}},
\end{equation}
and is jointly minimized during training. To analyze how different types of information contribute to the loss, we conceptually decompose the latent vectors $z_S$ and $z_T$, which are the outputs of the style module, into two parts:
\begin{equation}
z_S = \bigl[z_S^{\mathrm{r}},\; z_S^{\mathrm{i}}\bigr], 
\quad
z_T = \bigl[z_T^{\mathrm{r}},\; z_T^{\mathrm{i}}\bigr],
\end{equation}
where $z_S^{\mathrm{r}}$ and $z_T^{\mathrm{r}}$ encode disease-relevant features, while $z_S^{\mathrm{i}}$ and $z_T^{\mathrm{i}}$ capture irrelevant variations. This interpretation is used to characterize the distinct roles of latent subspaces during optimization. As shown in Appendix~\ref{app:style-proof}, $L_{\text{style}}$ penalizes all deviations from the prior, regardless of their semantic content. However, only components that also reduce $L_{\text{cluster}}$ by aligning the soft assignments across modalities contribute to lowering the total loss. This imbalance between the two loss terms naturally encourages the model to retain only informative features. Specifically, the disease-irrelevant components $z_S^{\mathrm{i}}$ and $z_T^{\mathrm{i}}$ often reflect factors such as differences in tissue structure that do not correlate with pathological conditions or fluctuations in the expression of non-cancer-related genes. These factors are inherently variable and lack consistency. As a result, they increase $L_{\text{style}}$ but fail to reduce $L_{\text{cluster}}$. Their independence across modalities implies that the prototype assignment distributions derived from these factors contain no mutual information and therefore almost never coincide. Since the Jensen–Shannon divergence in Eq.~\eqref{eq:L_cluster} is equal to zero only when the two distributions are identical, it remains strictly positive and cannot be reduced by modality-specific noise. Consequently, such irrelevant components lead to a higher total loss and are gradually suppressed during training. In contrast, the disease-relevant components $z_S^{\mathrm{r}}$ and $z_T^{\mathrm{r}}$ may capture features such as pathologically significant phenotypes in histopathology or cancer-associated gene expressions in transcriptomics. Although they may also incur a style penalty, they contribute to reducing $L_{\text{cluster}}$ by improving cross-modal consistency. Since they contribute to minimizing the total loss, these features are preserved.

In summary, the style clustering module effectively disentangles disease-relevant information from irrelevant variability, ensuring that only meaningful features are learned across modalities.

\subsection{Global Optimization Objective}

The global optimization objective is formally defined as:
\begin{equation} 
L = \lambda_{\alpha}L_{\text{align}} + \lambda_{\beta}L_{\text{retention}} + \lambda_{\gamma}L_{\text{style clustering}},
\label{eq:total_loss}
\end{equation}
where $\lambda_{\alpha}$, $\lambda_{\beta}$, and $\lambda_{\gamma}$ are hyperparameters balancing each term. Minimizing this composite loss function promotes the enrichment of disease-relevant information, capturing both modality-shared and modality-specific components, denoted by $S_{r,s}$, $S_{r,u}$, $T_{r,s}$, and $T_{r,u}$. Simultaneously, it suppresses disease-irrelevant information, represented by $S_{i,s}$, $S_{i,u}$, $T_{i,s}$, and $T_{i,u}$. This selective optimization ensures that the learned feature representations concentrate on patterns essential for accurate disease characterization across modalities, thereby improving the model's diagnostic and prognostic effectiveness.

\begin{table*}[!t]
\centering
\caption{Hyperparameter configurations used during pretraining and downstream task evaluations.}
\label{tab:hyperparams}
\begin{tabular}{l|c cc cc}
\toprule
\multirow{2}{*}{\textbf{Parameter}} 
& \multirow{2}{*}{\textbf{Pretraining}} 
& \multicolumn{2}{c}{\textbf{Subtyping}} 
& \multicolumn{2}{c}{\textbf{Survival}} \\
\cmidrule(lr){3-4} \cmidrule(lr){5-6}
& & \textbf{10-shot} & \textbf{All Data} & \textbf{10-shot} & \textbf{All Data} \\
\midrule
Optimizer & Adam & \multicolumn{2}{c}{Adam} & \multicolumn{2}{c}{Adam} \\
Learning rate & $2 \times 10^{-5}$ & \multicolumn{2}{c}{$2 \times 10^{-3}$} & \multicolumn{2}{c}{$3 \times 10^{-3}$} \\
Batch size & 16 & \multicolumn{2}{c}{16} & \multicolumn{2}{c}{16} \\
Epochs & 100 & \multicolumn{2}{c}{100} & \multicolumn{2}{c}{100} \\
Loss function & \proposedmethod~Loss & \multicolumn{2}{c}{Cross-entropy Loss} & \multicolumn{2}{c}{\makecell{Negative Log-Likelihood \\ Survival Loss}} \\
Slide mask ratio & 0.75 & \multicolumn{2}{c}{\textemdash} & \multicolumn{2}{c}{\textemdash} \\
Transcripts mask ratio & 0.75 & \multicolumn{2}{c}{\textemdash} & \multicolumn{2}{c}{\textemdash} \\
Mask strategy & Random & \multicolumn{2}{c}{\textemdash} & \multicolumn{2}{c}{\textemdash} \\
Initial $\tau$ value & $\exp(\log(1/0.07)) \approx 14.29$ & \multicolumn{2}{c}{\textemdash} & \multicolumn{2}{c}{\textemdash} \\
$L_{\text{align}}$ weight ($\lambda_{\alpha}$) & 0.5 & \multicolumn{2}{c}{\textemdash} & \multicolumn{2}{c}{\textemdash} \\
$L_{\text{retention}}$ weight ($\lambda_{\beta}$) & 0.3 & \multicolumn{2}{c}{\textemdash} & \multicolumn{2}{c}{\textemdash} \\
$L_{\text{style clustering}}$ weight ($\lambda_{\gamma}$) & 0.2 & \multicolumn{2}{c}{\textemdash} & \multicolumn{2}{c}{\textemdash} \\
Number of prototypes & 3{,}000 & \multicolumn{2}{c}{\textemdash} & \multicolumn{2}{c}{\textemdash} \\
Frozen encoder & \textemdash & \multicolumn{2}{c}{Yes} & \multicolumn{2}{c}{Yes} \\
\bottomrule
\end{tabular}
\end{table*}

\begin{table}[!t]
\centering
\renewcommand{\arraystretch}{1.2} %
\setlength{\tabcolsep}{8pt} %
\caption{Metadata of TCGA cohorts.}
\resizebox{\columnwidth}{!}{%
\begin{tabular}{@{}lcc@{}}
\toprule
\textbf{Cohort} & \textbf{Subtype} & \textbf{Total Samples} \\
\midrule
\multirow{2}{*}{\textbf{TCGA-BRCA}} & Invasive Ductal Carcinoma (IDC) & \multirow{2}{*}{\textbf{955}} \\
 & Invasive Lobular Carcinoma (ILC) & \\
\midrule
\multirow{2}{*}{\textbf{TCGA-NSCLC}} & Lung Adenocarcinoma (LUAD) & \multirow{2}{*}{\textbf{1,053}} \\
 & Lung Squamous Cell Carcinoma (LUSC) & \\
\midrule
\multirow{3}{*}{\textbf{TCGA-RCC}} & Kidney Renal Clear Cell Carcinoma (KIRC) & \multirow{3}{*}{\textbf{943}} \\
 & Kidney Renal Papillary Cell Carcinoma (KIRP) & \\
 & Kidney Chromophobe (KICH) & \\
\midrule
\multirow{2}{*}{\textbf{TCGA-COADREAD}} & Colon Adenocarcinoma (COAD) & \multirow{2}{*}{\textbf{623}} \\
 & Rectum Adenocarcinoma (READ) & \\
\bottomrule
\end{tabular}%
}
\label{tab:tcga_cohorts}
\end{table}

\section{Experiments and Results}
\subsection{Datasets}
Four distinct cohorts from the publicly available TCGA dataset~\cite{tomczak2015review} were utilized for evaluation: Breast Invasive Carcinoma (BRCA), Non-Small Cell Lung Cancer (NSCLC), Renal Cell Carcinoma (RCC), and Colon and Rectal Adenocarcinoma (COADREAD). Each cohort comprises specific subtypes, as detailed in Table~\ref{tab:tcga_cohorts}.

WSIs obtained from these cohorts were processed into patches at $20\times$ magnification, and feature representations were extracted using ResNet-50~\cite{he2016deep} and Phikon~\cite{filiot2023scaling}. Corresponding transcriptomics data were collected from Xena~\cite{goldman2020visualizing} and preprocessed using the proposed novel pipeline, reducing the original 198,620 genes to 40,146 for the BRCA cohort, 10,234 for the NSCLC cohort, 10,303 for the RCC cohort, and 20,056 for the COADREAD cohort, forming the novel refined transcriptomics datasets.

\begin{table*}[!t]
\centering
\renewcommand{\arraystretch}{1.2}  %
\setlength{\tabcolsep}{2pt}       %
\caption{Cancer subtype classification on TCGA-BRCA, TCGA-NSCLC, TCGA-RCC, and TCGA-COADREAD.}
\resizebox{\textwidth}{!}{
\begin{tabular}{@{}c c c | c c c c c c c c c c c c c c@{}}
\toprule
\multirow{2}{*}{\textbf{Dataset}} 
 & \multirow{2}{*}{\textbf{Backbone}} 
 & \multirow{2}{*}{\textbf{Setting}}
 & \multicolumn{2}{c}{ABMIL}
 & \multicolumn{2}{c}{ILRA-MIL}
 & \multicolumn{2}{c}{PORPOISE}
 & \multicolumn{2}{c}{SurvPath}
 & \multicolumn{2}{c}{Linear Classifier}
 & \multicolumn{2}{c}{TANGLE}
 & \multicolumn{2}{c}{\textbf{MIRROR}} \\
\cmidrule(lr){4-5} \cmidrule(lr){6-7} \cmidrule(lr){8-9} \cmidrule(lr){10-11} \cmidrule(lr){12-13} \cmidrule(lr){14-15} \cmidrule(lr){16-17}
& & & \textit{Acc} & \textit{F1}
      & \textit{Acc} & \textit{F1}
      & \textit{Acc} & \textit{F1}
      & \textit{Acc} & \textit{F1}
      & \textit{Acc} & \textit{F1}
      & \textit{Acc} & \textit{F1}
      & \textit{Acc} & \textit{F1} \\
\midrule

\multirow{8}{*}{\textbf{\begin{tabular}[c]{@{}c@{}}TCGA-\\NSCLC\end{tabular}}}
  & \multirow{4}{*}{ResNet-50}
    & \multirow{2}{*}{\textit{10-shot}}
    & 0.633 & 0.572
    & 0.634 & 0.596
    & 0.742 & 0.699
    & 0.654 & 0.618
    & 0.660 & 0.658
    & \underline{0.909} & \underline{0.909}
    & \textbf{0.930} & \textbf{0.930} \\
  & 
    & 
    & \scriptsize($\pm 0.084$) & \scriptsize($\pm 0.147$)
    & \scriptsize($\pm 0.072$) & \scriptsize($\pm 0.146$)
    & \scriptsize($\pm 0.182$) & \scriptsize($\pm 0.242$)
    & \scriptsize($\pm 0.049$) & \scriptsize($\pm 0.114$)
    & \scriptsize($\pm 0.024$) & \scriptsize($\pm 0.025$)
    & \scriptsize($\pm 0.033$) & \scriptsize($\pm 0.033$)
    & \scriptsize($\pm 0.006$) & \scriptsize($\pm 0.006$) \\
  & 
    & \multirow{2}{*}{\textit{All Data}}
    & 0.870 & 0.868
    & 0.890 & 0.886
    & 0.986 & 0.986
    & 0.967 & 0.966
    & 0.845 & 0.843
    & \underline{0.989} & \underline{0.987}
    & \textbf{0.992} & \textbf{0.992} \\
  & 
    & 
    & \scriptsize($\pm 0.015$) & \scriptsize($\pm 0.016$)
    & \scriptsize($\pm 0.014$) & \scriptsize($\pm 0.015$)
    & \scriptsize($\pm 0.007$) & \scriptsize($\pm 0.007$)
    & \scriptsize($\pm 0.027$) & \scriptsize($\pm 0.027$)
    & \scriptsize($\pm 0.014$) & \scriptsize($\pm 0.015$)
    & \scriptsize($\pm 0.009$) & \scriptsize($\pm 0.009$)
    & \scriptsize($\pm 0.011$) & \scriptsize($\pm 0.011$) \\
  & \multirow{4}{*}{Phikon}
    & \multirow{2}{*}{\textit{10-shot}}
    & 0.709 & 0.703
    & 0.682 & 0.678
    & 0.786 & 0.767
    & 0.739 & 0.670
    & 0.699 & 0.694
    & \underline{0.925} & \underline{0.924}
    & \textbf{0.939} & \textbf{0.938} \\
  & 
    & 
    & \scriptsize($\pm 0.030$) & \scriptsize($\pm 0.029$)
    & \scriptsize($\pm 0.043$) & \scriptsize($\pm 0.046$)
    & \scriptsize($\pm 0.155$) & \scriptsize($\pm 0.183$)
    & \scriptsize($\pm 0.060$) & \scriptsize($\pm 0.079$)
    & \scriptsize($\pm 0.030$) & \scriptsize($\pm 0.029$)
    & \scriptsize($\pm 0.036$) & \scriptsize($\pm 0.039$)
    & \scriptsize($\pm 0.017$) & \scriptsize($\pm 0.017$) \\
  & 
    & \multirow{2}{*}{\textit{All Data}}
    & 0.906 & 0.905
    & 0.917 & 0.916
    & \underline{0.984} & \underline{0.992}
    & 0.907 & 0.946
    & 0.901 & 0.900
    & 0.982 & 0.982
    & \textbf{0.994} & \textbf{0.994} \\
  & 
    & 
    & \scriptsize($\pm 0.010$) & \scriptsize($\pm 0.010$)
    & \scriptsize($\pm 0.017$) & \scriptsize($\pm 0.017$)
    & \scriptsize($\pm 0.007$) & \scriptsize($\pm 0.008$)
    & \scriptsize($\pm 0.023$) & \scriptsize($\pm 0.022$)
    & \scriptsize($\pm 0.008$) & \scriptsize($\pm 0.008$)
    & \scriptsize($\pm 0.012$) & \scriptsize($\pm 0.012$)
    & \scriptsize($\pm 0.007$) & \scriptsize($\pm 0.007$) \\

\midrule

\multirow{8}{*}{\textbf{\begin{tabular}[c]{@{}c@{}}TCGA-\\BRCA\end{tabular}}}
  & \multirow{4}{*}{ResNet-50}
    & \multirow{2}{*}{\textit{10-shot}}
    & 0.796 & 0.480
    & 0.714 & 0.402
    & \underline{0.902} & 0.726
    & 0.860 & 0.474
    & 0.862 & 0.507
    & \textbf{0.910} & \underline{0.757}
    & \textbf{0.910} & \textbf{0.786} \\
  & 
    & 
    & \scriptsize($\pm 0.075$) & \scriptsize($\pm 0.028$)
    & \scriptsize($\pm 0.322$) & \scriptsize($\pm 0.157$)
    & \scriptsize($\pm 0.010$) & \scriptsize($\pm 0.152$)
    & \scriptsize($\pm 0.031$) & \scriptsize($\pm 0.014$)
    & \scriptsize($\pm 0.031$) & \scriptsize($\pm 0.076$)
    & \scriptsize($\pm 0.036$) & \scriptsize($\pm 0.121$)
    & \scriptsize($\pm 0.032$) & \scriptsize($\pm 0.092$) \\
  & 
    & \multirow{2}{*}{\textit{All Data}}
    & 0.860 & 0.462
    & 0.926 & 0.831
    & 0.923 & 0.894
    & 0.911 & 0.862
    & 0.901 & 0.762
    & \underline{0.956} & \textbf{0.909}
    & \textbf{0.958} & \underline{0.902} \\
  & 
    & 
    & \scriptsize($\pm 0.030$) & \scriptsize($\pm 0.001$)
    & \scriptsize($\pm 0.016$) & \scriptsize($\pm 0.035$)
    & \scriptsize($\pm 0.011$) & \scriptsize($\pm 0.022$)
    & \scriptsize($\pm 0.030$) & \scriptsize($\pm 0.008$)
    & \scriptsize($\pm 0.029$) & \scriptsize($\pm 0.020$)
    & \scriptsize($\pm 0.018$) & \scriptsize($\pm 0.029$)
    & \scriptsize($\pm 0.017$) & \scriptsize($\pm 0.039$) \\
  & \multirow{4}{*}{Phikon}
    & \multirow{2}{*}{\textit{10-shot}}
    & 0.712 & 0.589
    & 0.807 & 0.566
    & 0.902 & 0.731
    & 0.876 & 0.671
    & 0.758 & 0.613
    & \underline{0.921} & \underline{0.829}
    & \textbf{0.923} & \textbf{0.832} \\
  & 
    & 
    & \scriptsize($\pm 0.131$) & \scriptsize($\pm 0.078$)
    & \scriptsize($\pm 0.040$) & \scriptsize($\pm 0.060$)
    & \scriptsize($\pm 0.030$) & \scriptsize($\pm 0.155$)
    & \scriptsize($\pm 0.111$) & \scriptsize($\pm 0.074$)
    & \scriptsize($\pm 0.059$) & \scriptsize($\pm 0.065$)
    & \scriptsize($\pm 0.031$) & \scriptsize($\pm 0.068$)
    & \scriptsize($\pm 0.025$) & \scriptsize($\pm 0.051$) \\
  & 
    & \multirow{2}{*}{\textit{All Data}}
    & 0.919 & 0.825
    & 0.926 & 0.848
    & 0.934 & \underline{0.899}
    & 0.890 & 0.862
    & 0.924 & 0.821
    & \underline{0.952} & 0.887
    & \textbf{0.955} & \textbf{0.902} \\
  & 
    & 
    & \scriptsize($\pm 0.008$) & \scriptsize($\pm 0.021$)
    & \scriptsize($\pm 0.025$) & \scriptsize($\pm 0.030$)
    & \scriptsize($\pm 0.030$) & \scriptsize($\pm 0.009$)
    & \scriptsize($\pm 0.030$) & \scriptsize($\pm 0.009$)
    & \scriptsize($\pm 0.038$) & \scriptsize($\pm 0.092$)
    & \scriptsize($\pm 0.022$) & \scriptsize($\pm 0.063$)
    & \scriptsize($\pm 0.023$) & \scriptsize($\pm 0.047$) \\

\midrule

\multirow{8}{*}{\textbf{\begin{tabular}[c]{@{}c@{}}TCGA-\\RCC\end{tabular}}}
  & \multirow{4}{*}{ResNet-50}
    & \multirow{2}{*}{\textit{10-shot}}
    & 0.752 & 0.629
    & 0.769 & 0.659
    & 0.937 & \underline{0.846}
    & 0.801 & 0.813
    & 0.752 & 0.673
    & \underline{0.938} & 0.832
    & \textbf{0.942} & \textbf{0.853} \\
  & 
    & 
    & \scriptsize($\pm 0.059$) & \scriptsize($\pm 0.121$)
    & \scriptsize($\pm 0.020$) & \scriptsize($\pm 0.104$)
    & \scriptsize($\pm 0.043$) & \scriptsize($\pm 0.112$)
    & \scriptsize($\pm 0.042$) & \scriptsize($\pm 0.090$)
    & \scriptsize($\pm 0.016$) & \scriptsize($\pm 0.099$)
    & \scriptsize($\pm 0.034$) & \scriptsize($\pm 0.111$)
    & \scriptsize($\pm 0.028$) & \scriptsize($\pm 0.126$) \\
  & 
    & \multirow{2}{*}{\textit{All Data}}
    & 0.908 & 0.677
    & 0.954 & 0.872
    & 0.976 & 0.857
    & 0.943 & \underline{0.907}
    & 0.933 & 0.827
    & \underline{0.988} & 0.903
    & \textbf{0.998} & \textbf{0.932} \\
  & 
    & 
    & \scriptsize($\pm 0.017$) & \scriptsize($\pm 0.144$)
    & \scriptsize($\pm 0.011$) & \scriptsize($\pm 0.139$)
    & \scriptsize($\pm 0.031$) & \scriptsize($\pm 0.194$)
    & \scriptsize($\pm 0.062$) & \scriptsize($\pm 0.186$)
    & \scriptsize($\pm 0.007$) & \scriptsize($\pm 0.116$)
    & \scriptsize($\pm 0.007$) & \scriptsize($\pm 0.138$)
    & \scriptsize($\pm 0.003$) & \scriptsize($\pm 0.150$) \\
  & \multirow{4}{*}{Phikon}
    & \multirow{2}{*}{\textit{10-shot}}
    & 0.839 & 0.733
    & 0.845 & 0.751
    & \underline{0.951} & 0.854
    & 0.924 & 0.832
    & 0.828 & 0.727
    & 0.950 & \underline{0.867}
    & \textbf{0.952} & \textbf{0.931} \\
  & 
    & 
    & \scriptsize($\pm 0.029$) & \scriptsize($\pm 0.083$)
    & \scriptsize($\pm 0.021$) & \scriptsize($\pm 0.090$)
    & \scriptsize($\pm 0.030$) & \scriptsize($\pm 0.114$)
    & \scriptsize($\pm 0.110$) & \scriptsize($\pm 0.061$)
    & \scriptsize($\pm 0.027$) & \scriptsize($\pm 0.107$)
    & \scriptsize($\pm 0.014$) & \scriptsize($\pm 0.123$)
    & \scriptsize($\pm 0.030$) & \scriptsize($\pm 0.042$) \\
  & 
    & \multirow{2}{*}{\textit{All Data}}
    & 0.968 & 0.886
    & 0.976 & 0.902
    & 0.988 & 0.920
    & 0.977 & 0.873
    & 0.961 & 0.885
    & \underline{0.989} & \underline{0.923}
    & \textbf{0.997} & \textbf{0.931} \\
  & 
    & 
    & \scriptsize($\pm 0.015$) & \scriptsize($\pm 0.133$)
    & \scriptsize($\pm 0.007$) & \scriptsize($\pm 0.016$)
    & \scriptsize($\pm 0.011$) & \scriptsize($\pm 0.015$)
    & \scriptsize($\pm 0.029$) & \scriptsize($\pm 0.027$)
    & \scriptsize($\pm 0.011$) & \scriptsize($\pm 0.130$)
    & \scriptsize($\pm 0.001$) & \scriptsize($\pm 0.147$)
    & \scriptsize($\pm 0.003$) & \scriptsize($\pm 0.149$) \\

\midrule

\multirow{8}{*}{\textbf{\begin{tabular}[c]{@{}c@{}}TCGA-\\COADREAD\end{tabular}}}
  & \multirow{4}{*}{ResNet-50}
    & \multirow{2}{*}{\textit{10-shot}}
    & 0.815 & \underline{0.542}
    & 0.781 & 0.461
    & 0.813 & 0.448
    & 0.820 & 0.476
    & 0.819 & 0.502
    & \underline{0.831} & 0.539
    & \textbf{0.834} & \textbf{0.556} \\
  & 
    & 
    & \scriptsize($\pm 0.057$) & \scriptsize($\pm 0.066$)
    & \scriptsize($\pm 0.102$) & \scriptsize($\pm 0.034$)
    & \scriptsize($\pm 0.065$) & \scriptsize($\pm 0.020$)
    & \scriptsize($\pm 0.055$) & \scriptsize($\pm 0.049$)
    & \scriptsize($\pm 0.055$) & \scriptsize($\pm 0.080$)
    & \scriptsize($\pm 0.043$) & \scriptsize($\pm 0.066$)
    & \scriptsize($\pm 0.039$) & \scriptsize($\pm 0.144$) \\
  & 
    & \multirow{2}{*}{\textit{All Data}}
    & 0.828 & 0.515
    & 0.813 & 0.448
    & 0.814 & 0.447
    & 0.830 & 0.523
    & 0.816 & 0.470
    & \underline{0.928} & \underline{0.881}
    & \textbf{0.938} & \textbf{0.893} \\
  & 
    & 
    & \scriptsize($\pm 0.051$) & \scriptsize($\pm 0.091$)
    & \scriptsize($\pm 0.065$) & \scriptsize($\pm 0.020$)
    & \scriptsize($\pm 0.065$) & \scriptsize($\pm 0.020$)
    & \scriptsize($\pm 0.065$) & \scriptsize($\pm 0.020$)
    & \scriptsize($\pm 0.069$) & \scriptsize($\pm 0.064$)
    & \scriptsize($\pm 0.042$) & \scriptsize($\pm 0.051$)
    & \scriptsize($\pm 0.022$) & \scriptsize($\pm 0.017$) \\
  & \multirow{4}{*}{Phikon}
    & \multirow{2}{*}{\textit{10-shot}}
    & 0.659 & 0.495
    & 0.691 & \underline{0.512}
    & \underline{0.813} & 0.448
    & 0.812 & 0.446
    & 0.644 & 0.510
    & 0.772 & 0.483
    & \textbf{0.828} & \textbf{0.515} \\
  & 
    & 
    & \scriptsize($\pm 0.133$) & \scriptsize($\pm 0.099$)
    & \scriptsize($\pm 0.102$) & \scriptsize($\pm 0.068$)
    & \scriptsize($\pm 0.065$) & \scriptsize($\pm 0.020$)
    & \scriptsize($\pm 0.065$) & \scriptsize($\pm 0.020$)
    & \scriptsize($\pm 0.118$) & \scriptsize($\pm 0.073$)
    & \scriptsize($\pm 0.087$) & \scriptsize($\pm 0.103$)
    & \scriptsize($\pm 0.043$) & \scriptsize($\pm 0.101$) \\
  & 
    & \multirow{2}{*}{\textit{All Data}}
    & 0.834 & 0.554
    & 0.813 & 0.548
    & 0.859 & 0.552
    & 0.815 & 0.449
    & 0.819 & 0.570
    & \underline{0.928} & \underline{0.880}
    & \textbf{0.941} & \textbf{0.902} \\
  & 
    & 
    & \scriptsize($\pm 0.037$) & \scriptsize($\pm 0.114$)
    & \scriptsize($\pm 0.065$) & \scriptsize($\pm 0.020$)
    & \scriptsize($\pm 0.065$) & \scriptsize($\pm 0.217$)
    & \scriptsize($\pm 0.065$) & \scriptsize($\pm 0.020$)
    & \scriptsize($\pm 0.069$) & \scriptsize($\pm 0.096$)
    & \scriptsize($\pm 0.049$) & \scriptsize($\pm 0.046$)
    & \scriptsize($\pm 0.020$) & \scriptsize($\pm 0.018$) \\

\bottomrule
\end{tabular}
}
\label{tab:subtyping}
\end{table*}

\subsection{Implementation Details}
\subsubsection{Data Preprocessing}
For WSIs, adhering to the conventional preprocessing protocol established in~\cite{campanella2019clinical}, the foreground tissue regions are first segmented using Otsu's method~\cite{otsu1975threshold}. These segmented regions are then divided into patches, forming an instance bag that is processed by a pre-trained patch encoder to extract feature representations. To enable batched training, we perform random sampling to select a fixed number of feature representations, employing replacement if the available number of representations is less than the required count.

For raw transcriptomics data, we first apply RFE with 5-fold cross-validation for each cohort to identify the most performant support set for the subtyping task. To enhance interpretability from a biological perspective, we manually incorporate genes associated with specific cancer subtypes based on the COSMIC database~\cite{sondka2024cosmic}, resulting in a one-dimensional transcriptomics feature vector.

\subsubsection{Experiment Setup}

All experiments, including both pretraining and downstream task evaluations, are conducted under a 5-fold cross-validation protocol. Each dataset is randomly partitioned into five folds. In each run, one fold is held out for evaluation while the remaining four folds are used for training. Specifically, under the 10-shot setting, training samples are randomly selected from the four training folds. The evaluation fold remains strictly held out and is never accessed during training, ensuring no data leakage. All split definitions and sampling scripts are released in our public code repository to support reproducibility.

Pretraining is performed separately for each cohort and repeated under each cross-validation split. This ensures that no evaluation samples are used during representation learning, maintaining a fair and unbiased evaluation setting.

After pretraining, \proposedmethod~is further evaluated on cancer subtyping and survival analysis tasks by concatenating the outputs from the two modality-specific encoders and feeding the combined vector into a linear classifier. We adopt both few-shot learning and linear probing for evaluation, following the same 5-fold cross-validation setup. To further assess generalization ability, we conduct evaluations using two different pretrained patch encoders.

The model is optimized using the Adam optimizer~\cite{kingma2014adam} and is implemented using PyTorch~\cite{paszke2019pytorch}. All training sessions are performed on a single NVIDIA GeForce RTX 3090 GPU paired with an Intel i7-12700K CPU and 32GB of memory. Full hyperparameter configurations for both pretraining and downstream task evaluations are provided in Table~\ref{tab:hyperparams}. All hyperparameter settings and training configurations are also released in our public code repository.

\subsection{Downstream Tasks} 
\label{subsec:downstream_tasks}
Two tasks are used for downstream evaluations, namely subtype classification and survival analysis. We compared the proposed \proposedmethod\ against ABMIL~\cite{ilse2018attention}, ILRA-MIL~\cite{xiang2023exploring}, PORPOISE~\cite{chen2022pan}, SurvPath~\cite{jaume2024modeling}, Linear Classifier, and TANGLE~\cite{jaume2024transcriptomics}, utilizing two distinct backbones under both 10-shot and linear probing settings. Among these methods, ABMIL, ILRA-MIL, and Linear Classifier take histopathology as input, whereas the other models integrate both histopathology and transcriptomics data. To ensure fair comparisons and optimal performance, all methods are trained on the proposed transcriptomic datasets. Notably, the original TANGLE framework evaluates only histopathology features. To comprehensively assess its performance and maintain methodological consistency, we extend TANGLE by incorporating both histopathology and transcriptomic features, aligning its input with that of \proposedmethod.

\subsubsection{Subtype Classification}
Accuracy and F1 score are employed as performance metrics to evaluate the classification effectiveness of the compared methods. 

As illustrated in Table~\ref{tab:subtyping}, the proposed \proposedmethod\ consistently achieves superior performance, demonstrating its effectiveness in integrating histopathology and transcriptomic data. On the TCGA-NSCLC dataset, \proposedmethod\ achieves the highest accuracy and F1 score across all settings, outperforming all baseline methods. Notably, its strong performance remains consistent regardless of the chosen backbone architecture, highlighting its robustness on the TCGA-NSCLC dataset. Similarly, on the TCGA-BRCA dataset, \proposedmethod\ achieves state-of-the-art results, surpassing the second best model's (TANGLE) result by 1.5\% in F1 score under the linear probing setting with the Phikon backbone. Additionally, \proposedmethod\ significantly outperforms TANGLE by 6.4\% in F1 score on the TCGA-RCC dataset under the 10-shot setting with the Phikon backbone. On the TCGA-COADREAD dataset, it surpasses TANGLE by 1.3\% in accuracy under the linear probing setting with the Phikon backbone. These findings validate the robustness and generalizability of \proposedmethod\ across different cancer subtypes and backbone architectures.

\begin{table*}[!ht]
\centering
\renewcommand{\arraystretch}{1.2} %
\setlength{\tabcolsep}{2pt} %
\caption{Cancer survival analysis on TCGA-BRCA and TCGA-NSCLC, TCGA-RCC, and TCGA-COADREAD.}
\begin{tabular}{@{}ccl|ccccccc@{}}
\toprule
\textbf{Dataset} & \textbf{Backbone} & \textbf{Setting} & ABMIL & ILRA-MIL & PORPOISE & SurvPath & Linear Classifier & TANGLE & \textbf{MIRROR} \\
\midrule
\multirow{4}{*}{\begin{tabular}[c]{@{}c@{}}\textbf{TCGA-}\\ \textbf{NSCLC}\end{tabular}}
& \multirow{2}{*}{ResNet-50} & \textit{10-shot}
& 0.529 \scriptsize($\pm 0.033$)
& 0.593 \scriptsize($\pm 0.045$)
& 0.565 \scriptsize($\pm 0.011$)
& 0.547 \scriptsize($\pm 0.071$)
& \underline{0.604} \scriptsize($\pm 0.019$)
& 0.570 \scriptsize($\pm 0.055$)
& \textbf{0.605} \scriptsize($\pm 0.011$) \\
& & \textit{All Data}
& 0.538 \scriptsize($\pm 0.042$)
& 0.606 \scriptsize($\pm 0.038$)
& 0.614 \scriptsize($\pm 0.047$)
& 0.613 \scriptsize($\pm 0.048$)
& \underline{0.618} \scriptsize($\pm 0.041$)
& 0.565 \scriptsize($\pm 0.042$)
& \textbf{0.621} \scriptsize($\pm 0.054$) \\

& \multirow{2}{*}{Phikon} & \textit{10-shot}
& 0.557 \scriptsize($\pm 0.016$)
& 0.574 \scriptsize($\pm 0.008$)
& 0.577 \scriptsize($\pm 0.022$)
& 0.573 \scriptsize($\pm 0.056$)
& 0.577 \scriptsize($\pm 0.017$)
& \underline{0.583} \scriptsize($\pm 0.049$)
& \textbf{0.584} \scriptsize($\pm 0.007$) \\
& & \textit{All Data}
& 0.567 \scriptsize($\pm 0.039$)
& \underline{0.608} \scriptsize($\pm 0.025$)
& 0.602 \scriptsize($\pm 0.053$)
& 0.572 \scriptsize($\pm 0.031$)
& 0.596 \scriptsize($\pm 0.035$)
& 0.593 \scriptsize($\pm 0.039$)
& \textbf{0.613} \scriptsize($\pm 0.043$) \\

\midrule
\multirow{4}{*}{\begin{tabular}[c]{@{}c@{}}\textbf{TCGA-}\\ \textbf{BRCA}\end{tabular}}
& \multirow{2}{*}{ResNet-50} & \textit{10-shot}
& 0.575 \scriptsize($\pm 0.054$)
& 0.587 \scriptsize($\pm 0.041$)
& 0.601 \scriptsize($\pm 0.028$)
& 0.560 \scriptsize($\pm 0.039$)
& 0.605 \scriptsize($\pm 0.044$)
& \underline{0.607} \scriptsize($\pm 0.052$)
& \textbf{0.612} \scriptsize($\pm 0.046$) \\
& & \textit{All Data}
& 0.573 \scriptsize($\pm 0.077$)
& 0.616 \scriptsize($\pm 0.026$)
& \underline{0.659} \scriptsize($\pm 0.091$)
& 0.651 \scriptsize($\pm 0.019$)
& 0.657 \scriptsize($\pm 0.029$)
& 0.580 \scriptsize($\pm 0.033$)
& \textbf{0.671} \scriptsize($\pm 0.096$) \\

& \multirow{2}{*}{Phikon} & \textit{10-shot}
& 0.544 \scriptsize($\pm 0.027$)
& \underline{0.616} \scriptsize($\pm 0.067$)
& 0.608 \scriptsize($\pm 0.031$)
& 0.603 \scriptsize($\pm 0.044$)
& 0.587 \scriptsize($\pm 0.042$)
& 0.602 \scriptsize($\pm 0.039$)
& \textbf{0.623} \scriptsize($\pm 0.054$) \\
& & \textit{All Data}
& 0.551 \scriptsize($\pm 0.029$)
& 0.645 \scriptsize($\pm 0.065$)
& 0.500 \scriptsize($\pm 0.068$)
& 0.650 \scriptsize($\pm 0.072$)
& \underline{0.658} \scriptsize($\pm 0.048$)
& 0.540 \scriptsize($\pm 0.055$)
& \textbf{0.665} \scriptsize($\pm 0.082$) \\

\midrule
\multirow{4}{*}{\begin{tabular}[c]{@{}c@{}}\textbf{TCGA-}\\ \textbf{RCC}\end{tabular}}
& \multirow{2}{*}{ResNet-50} & \textit{10-shot}
& 0.598 \scriptsize($\pm 0.047$)
& 0.647 \scriptsize($\pm 0.044$)
& \underline{0.669} \scriptsize($\pm 0.046$)
& 0.641 \scriptsize($\pm 0.040$)
& 0.651 \scriptsize($\pm 0.051$)
& 0.603 \scriptsize($\pm 0.072$)
& \textbf{0.697} \scriptsize($\pm 0.013$) \\
& & \textit{All Data}
& 0.596 \scriptsize($\pm 0.030$)
& 0.754 \scriptsize($\pm 0.048$)
& 0.777 \scriptsize($\pm 0.020$)
& 0.750 \scriptsize($\pm 0.040$)
& 0.743 \scriptsize($\pm 0.040$)
& \underline{0.794} \scriptsize($\pm 0.040$)
& \textbf{0.800} \scriptsize($\pm 0.045$) \\

& \multirow{2}{*}{Phikon} & \textit{10-shot}
& 0.503 \scriptsize($\pm 0.040$)
& 0.672 \scriptsize($\pm 0.041$)
& \underline{0.694} \scriptsize($\pm 0.022$)
& 0.630 \scriptsize($\pm 0.058$)
& 0.651 \scriptsize($\pm 0.056$)
& 0.677 \scriptsize($\pm 0.083$)
& \textbf{0.739} \scriptsize($\pm 0.032$) \\
& & \textit{All Data}
& 0.502 \scriptsize($\pm 0.041$)
& 0.783 \scriptsize($\pm 0.046$)
& 0.780 \scriptsize($\pm 0.039$)
& 0.756 \scriptsize($\pm 0.037$)
& 0.778 \scriptsize($\pm 0.051$)
& \underline{0.801} \scriptsize($\pm 0.044$)
& \textbf{0.803} \scriptsize($\pm 0.043$) \\

\midrule
\multirow{4}{*}{\begin{tabular}[c]{@{}c@{}}\textbf{TCGA-}\\ \textbf{COADREAD}\end{tabular}}
& \multirow{2}{*}{ResNet-50} & \textit{10-shot}
& 0.591 \scriptsize($\pm 0.048$)
& 0.569 \scriptsize($\pm 0.030$)
& \underline{0.598} \scriptsize($\pm 0.042$)
& 0.571 \scriptsize($\pm 0.063$)
& 0.595 \scriptsize($\pm 0.065$)
& 0.573 \scriptsize($\pm 0.071$)
& \textbf{0.619} \scriptsize($\pm 0.050$) \\
& & \textit{All Data}
& 0.601 \scriptsize($\pm 0.050$)
& 0.631 \scriptsize($\pm 0.057$)
& 0.616 \scriptsize($\pm 0.054$)
& 0.671 \scriptsize($\pm 0.093$)
& \underline{0.706} \scriptsize($\pm 0.054$)
& 0.574 \scriptsize($\pm 0.071$)
& \textbf{0.730} \scriptsize($\pm 0.057$) \\

& \multirow{2}{*}{Phikon} & \textit{10-shot}
& 0.490 \scriptsize($\pm 0.054$)
& 0.611 \scriptsize($\pm 0.075$)
& 0.601 \scriptsize($\pm 0.112$)
& 0.620 \scriptsize($\pm 0.066$)
& \underline{0.639} \scriptsize($\pm 0.053$)
& 0.630 \scriptsize($\pm 0.094$)
& \textbf{0.657} \scriptsize($\pm 0.054$) \\
& & \textit{All Data}
& 0.538 \scriptsize($\pm 0.021$)
& \underline{0.713} \scriptsize($\pm 0.024$)
& 0.642 \scriptsize($\pm 0.047$)
& 0.660 \scriptsize($\pm 0.068$)
& 0.701 \scriptsize($\pm 0.043$)
& 0.618 \scriptsize($\pm 0.076$)
& \textbf{0.721} \scriptsize($\pm 0.033$) \\

\bottomrule
\end{tabular}
\label{tab:survival}
\end{table*}

\subsubsection{Survival Prediction}
To assess the performance of our model in survival analysis, we employ a discrete survival model that categorizes patients into four distinct bins. The Concordance Index (C-index) is used as the primary metric to assess performance.

As shown in Table~\ref{tab:survival}, the proposed \proposedmethod\ achieves significant performance improvements over all baseline methods. On the TCGA-NSCLC dataset, \proposedmethod\ surpasses PORPOISE by $4.0\%$ and achieves a $7.6\%$ improvement compared to ABMIL under the 10-shot setting with ResNet-50 as the backbone. This enhancement highlights the superior capability of our model to leverage the integrated multimodal features effectively. Additionally, \proposedmethod\ consistently outperforms TANGLE across all settings, further emphasizing its robust feature representation and generalizability. These results highlight the effectiveness of \proposedmethod\ in tackling complex survival prediction tasks, reinforcing its potential for improving prognostic modeling in computational pathology.

\subsection{Ablation Study}

\subsubsection{Ablation Study of Model Components}
We conducted an ablation study to evaluate the effectiveness of the key components of \proposedmethod, including the modality alignment module, the modality retention module, and the style clustering module. The study was performed on the TCGA-NSCLC, TCGA-BRCA, TCGA-RCC, and TCGA-COADREAD datasets using linear probing with Phikon as the backbone, comparing \proposedmethod's performance both with and without these components.

As shown in Table~\ref{tab:ablation}, the results highlight the contributions of each module. The modality retention module enhances subtype classification, demonstrating its ability to preserve modality-specific information. However, its inclusion slightly decreases the performance of survival prediction. This reduction arises because the module retains not only disease-relevant features but also disease-irrelevant variations, such as staining artifacts or non-prognostic gene fluctuations. Survival prediction is particularly sensitive to such variance, as censoring and the resulting class imbalance reduce label informativeness and lower the effective signal-to-noise ratio compared with subtype classification. These nuisance variations can distort risk calibration and disrupt the relative ordering of patients, thereby depressing the C-index. The style clustering module effectively addresses this limitation by encouraging the model to emphasize disease-relevant patterns while suppressing redundancy, which yields consistent gains in survival prediction. While its inclusion results in a minor decline in subtype classification on TCGA-BRCA, performance remains stable on other datasets, and this small drawback is fully mitigated when both modules are jointly applied. Ultimately, \proposedmethod\ achieves the best overall performance, confirming the synergistic benefits of the proposed components.

\subsubsection{Ablation Study of Style Clustering Module}
To empirically validate the claims in Section~\ref{subsec:style_clustering}, we evaluated the individual and joint contributions of the style module and the clustering module. In the style ablation, the stochastic encoder was replaced with a deterministic MLP that produced a direct embedding $z$ instead of learning distributional parameters $\mu$ and $\log\sigma$. In the clustering ablation, the prototype-based soft alignment was replaced with a hard alignment between histopathology and transcriptomics embeddings, thereby enforcing a rigid matching strategy in place of probabilistic prototype consistency. All experiments employed Phikon as the backbone under the linear probing setting on TCGA-NSCLC, TCGA-BRCA, TCGA-RCC, and TCGA-COADREAD.

The results presented in Table~\ref{tab:ablation_style_clustering} demonstrate that the two modules function as complementary forces that jointly achieve compression and promotion of disease-relevant representations. The style loss enforces compactness by penalizing deviations from the standard normal prior, which suppresses redundant degrees of freedom and reduces variance. When used in isolation, this compression may suppress both disease-relevant and irrelevant information, thereby diminishing overall performance. In contrast, the clustering loss promotes disease-relevant alignment by encouraging convergence of the probabilistic assignments toward shared prototypes. Applied alone, this alignment can also admit nuisance variability, since there is no mechanism to constrain the latent space. When both modules are combined, compression removes redundant variation while alignment preserves and organizes task-relevant factors, resulting in consistent improvements across cohorts and tasks. These findings confirm the intended design of the style clustering module, in which the two objectives counterbalance each other and guide the representations toward a stable state that eliminates redundancy while retaining disease-relevant structure.

\begin{table*}[!t]
\centering
\renewcommand{\arraystretch}{1.2}
\setlength{\tabcolsep}{5pt}
\caption{Ablation study of model components.}
\begin{tabular}{@{}ccc|cc cc cc cc@{}}
\toprule
\multicolumn{3}{c|}{\textbf{Module Setting}} &
\multicolumn{2}{c}{\textbf{TCGA-NSCLC}} &
\multicolumn{2}{c}{\textbf{TCGA-BRCA}} &
\multicolumn{2}{c}{\textbf{TCGA-RCC}} &
\multicolumn{2}{c}{\textbf{TCGA-COADREAD}} \\
\cmidrule(lr){4-5}\cmidrule(lr){6-7}\cmidrule(lr){8-9}\cmidrule(lr){10-11}
\textit{Alignment} & \textit{Retention} & \textit{Style Clustering} &
\textbf{Subtyping} & \textbf{Survival} &
\textbf{Subtyping} & \textbf{Survival} &
\textbf{Subtyping} & \textbf{Survival} &
\textbf{Subtyping} & \textbf{Survival} \\
\midrule
\multirow{2}{*}{\checkmark} & \multirow{2}{*}{} & \multirow{2}{*}{} &
0.986 & 0.600 & 0.953 & 0.612 & 0.992 & 0.775 & 0.898 & 0.697 \\
& & &
\scriptsize($\pm\,0.009$) & \scriptsize($\pm\,0.040$) &
\scriptsize($\pm\,0.024$) & \scriptsize($\pm\,0.075$) &
\scriptsize($\pm\,0.003$) & \scriptsize($\pm\,0.051$) &
\scriptsize($\pm\,0.023$) & \scriptsize($\pm\,0.089$) \\
\multirow{2}{*}{\checkmark} & \multirow{2}{*}{\checkmark} & \multirow{2}{*}{} &
0.988 & 0.595 & 0.954 & 0.610 & 0.994 & 0.773 & 0.918 & 0.647 \\
& & &
\scriptsize($\pm\,0.012$) & \scriptsize($\pm\,0.024$) &
\scriptsize($\pm\,0.028$) & \scriptsize($\pm\,0.085$) &
\scriptsize($\pm\,0.003$) & \scriptsize($\pm\,0.040$) &
\scriptsize($\pm\,0.059$) & \scriptsize($\pm\,0.049$) \\
\multirow{2}{*}{\checkmark} & \multirow{2}{*}{} & \multirow{2}{*}{\checkmark} &
0.989 & 0.610 & 0.950 & 0.654 & 0.996 & 0.778 & 0.921 & 0.702 \\
& & &
\scriptsize($\pm\,0.003$) & \scriptsize($\pm\,0.044$) &
\scriptsize($\pm\,0.016$) & \scriptsize($\pm\,0.083$) &
\scriptsize($\pm\,0.003$) & \scriptsize($\pm\,0.045$) &
\scriptsize($\pm\,0.027$) & \scriptsize($\pm\,0.034$) \\
\multirow{2}{*}{\checkmark} & \multirow{2}{*}{\checkmark} & \multirow{2}{*}{\checkmark} &
\textbf{0.994} & \textbf{0.613} & \textbf{0.955} & \textbf{0.665} & \textbf{0.997} & \textbf{0.803} & \textbf{0.941} & \textbf{0.721} \\
& & &
\scriptsize($\pm\,0.007$) & \scriptsize($\pm\,0.043$) &
\scriptsize($\pm\,0.023$) & \scriptsize($\pm\,0.082$) &
\scriptsize($\pm\,0.003$) & \scriptsize($\pm\,0.043$) &
\scriptsize($\pm\,0.020$) & \scriptsize($\pm\,0.033$) \\
\bottomrule
\end{tabular}%
\label{tab:ablation}
\end{table*}

\subsubsection{Ablation Study of Gene Preprocessing Methods}
We compared our preprocessing pipeline with two widely used alternatives. The first is the preprocessed transcriptomics data from PORPOISE~\cite{chen2022pan}, which collects raw data from cBioPortal~\cite{cerami2012cbio} and reduces the dimensionality of the collected RNA sequencing features by selecting gene sets from the gene family categories in the Molecular Signatures Database (MSigDB)~\cite{liberzon2015molecular}. The second is the preprocessing adopted by SurvPath~\cite{jaume2024modeling}, where raw transcriptomic profiles are obtained from the Xena~\cite{goldman2020visualizing} and pathways are constructed from Reactome~\cite{gillespie2022reactome} and the MSigDB Hallmarks collection. Although SurvPath operates at the pathway rather than the gene level, we include it here for completeness. Since neither PORPOISE nor SurvPath has released their preprocessing code, and thus our comparison relies on the processed data they provide, with TCGA-BRCA being the only cohort aligned with our setting. All experiments employed Phikon as the backbone under the linear probing setting on TCGA-BRCA. As summarized in Table~\ref{tab:ablation_gene_preproc}, our preprocessing approach consistently achieves the best overall performance across both subtyping and survival tasks, outperforming PORPOISE and SurvPath. These results confirm the effectiveness of our preprocessing strategy.

\subsubsection{Ablation Study of Loss Term Weights}
To determine the hyperparameters in Eq.~\eqref{eq:total_loss}, we conducted an ablation study on the loss term weights. The evaluation employed Phikon as the backbone for feature extraction under the linear probing setting on TCGA-BRCA and followed the same 5-fold cross-validation protocol. As reported in Table~\ref{tab:ablation_loss}, we varied the relative contributions of $\lambda_{\alpha}$, $\lambda_{\beta}$, and $\lambda_{\gamma}$, which respectively balance the alignment loss $L_{\text{align}}$, the retention loss $L_{\text{retention}}$, and the style clustering loss $L_{\text{style clustering}}$, while keeping their sum equal to one. The results indicate that performance remains stable across different configurations, with all three settings producing comparable outcomes. The configuration $\lambda_{\alpha}=0.5$, $\lambda_{\beta}=0.3$, and $\lambda_{\gamma}=0.2$ provided the best overall balance, delivering the highest accuracy for subtype classification and the strongest C-index for survival prediction. Based on these findings, we adopt this configuration throughout the manuscript. This analysis demonstrates that our method is not overly sensitive to the precise choice of hyperparameters and that the reported improvements reflect robust performance rather than fragile tuning.

\subsection{Qualitative Analysis}

Here, we demonstrate the explainability of \proposedmethod\ with attention weights visualization from the slide encoder and UMAP~\cite{mcinnes2018umap} projections of features extracted by slide and RNA encoders.

\begin{figure*}[!t]
\centering
\includegraphics[width=\textwidth]{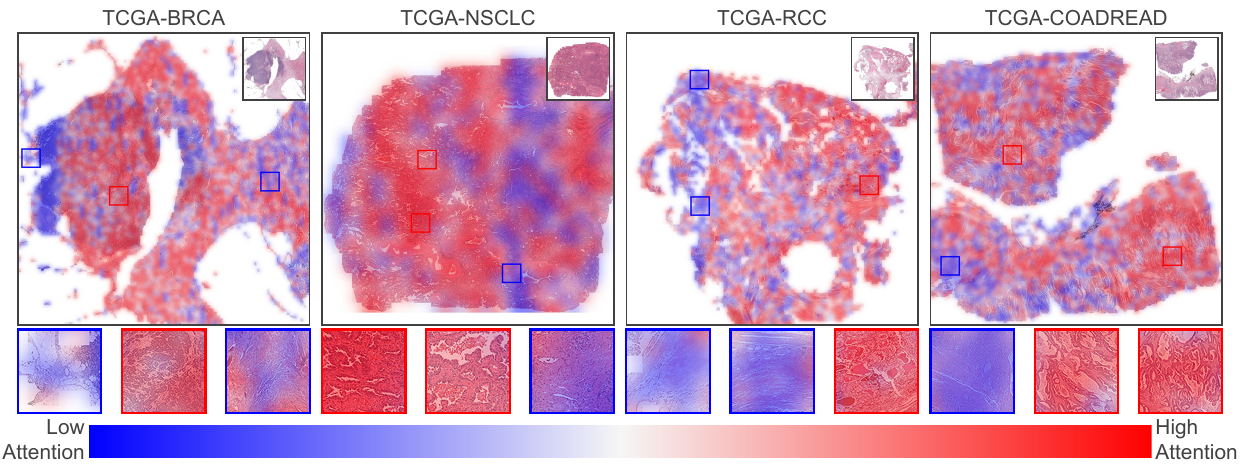}
\caption{\textbf{Visualization of slide encoder attention weights on TCGA-BRCA, TCGA-NSCLC, TCGA-RCC and TCGA-COADREAD.} Regions exhibiting higher attention scores predominantly correspond to malignant, tumor-bearing tissue, whereas areas with lower scores typically indicate normal regions.}
\label{fig:wsi_attention_map}
\end{figure*}

As illustrated in Figure~\ref{fig:wsi_attention_map}, we selected four representative samples from each cohort used. The attention weights from the slide encoder are visualized using heatmaps to highlight regions of interest. Areas shown in warmer colors indicate regions that the model pays more attention to. The results suggest \proposedmethod\ consistently attends to disease-relevant regions, particularly to cancerous regions. This strong alignment showcases \proposedmethod's extraordinary interpretability, confirming that it not only excels in downstream tasks but also provides biologically grounded decision-making insights. Across all four TCGA cohorts, regions receiving higher attention consistently correspond to cancerous tissue characterized by dense malignant cellularity and invasive fronts, whereas regions with lower attention are predominantly associated with non-neoplastic compartments, including background stroma, adipose tissue, and non-tissue artifacts. These consistent patterns offer a clear biological interpretation of both high- and low-attention patches.

\begin{table*}[!t]
\centering
\renewcommand{\arraystretch}{1.2}
\setlength{\tabcolsep}{5pt}
\caption{Ablation study of style clustering module.}
\begin{tabular}{@{}cc|cc cc cc cc@{}}
\toprule
\multicolumn{2}{c|}{\textbf{Module Setting}} &
\multicolumn{2}{c}{\textbf{TCGA-NSCLC}} &
\multicolumn{2}{c}{\textbf{TCGA-BRCA}} &
\multicolumn{2}{c}{\textbf{TCGA-RCC}} &
\multicolumn{2}{c}{\textbf{TCGA-COADREAD}} \\
\cmidrule(lr){3-4}\cmidrule(lr){5-6}\cmidrule(lr){7-8}\cmidrule(lr){9-10}
\textit{Style} & \textit{Clustering} &
\textbf{Subtyping} & \textbf{Survival} &
\textbf{Subtyping} & \textbf{Survival} &
\textbf{Subtyping} & \textbf{Survival} &
\textbf{Subtyping} & \textbf{Survival} \\
\midrule
\multirow{2}{*}{\checkmark} & \multirow{2}{*}{} &
0.984 & 0.596 & 0.939 & 0.625 & 0.993 & 0.772 & 0.887 & 0.636 \\
& &
\scriptsize($\pm\,0.007$) & \scriptsize($\pm\,0.023$) &
\scriptsize($\pm\,0.026$) & \scriptsize($\pm\,0.035$) &
\scriptsize($\pm\,0.004$) & \scriptsize($\pm\,0.051$) &
\scriptsize($\pm\,0.060$) & \scriptsize($\pm\,0.039$) \\
\multirow{2}{*}{} & \multirow{2}{*}{\checkmark} &
0.985 & 0.598 & 0.945 & 0.634 & 0.994 & 0.768 & 0.896 & 0.633 \\
& &
\scriptsize($\pm\,0.006$) & \scriptsize($\pm\,0.033$) &
\scriptsize($\pm\,0.021$) & \scriptsize($\pm\,0.044$) &
\scriptsize($\pm\,0.004$) & \scriptsize($\pm\,0.033$) &
\scriptsize($\pm\,0.024$) & \scriptsize($\pm\,0.048$) \\
\multirow{2}{*}{\checkmark} & \multirow{2}{*}{\checkmark} &
\textbf{0.994} & \textbf{0.613} & \textbf{0.955} & \textbf{0.665} & \textbf{0.997} & \textbf{0.803} & \textbf{0.941} & \textbf{0.721} \\
& &
\scriptsize($\pm\,0.007$) & \scriptsize($\pm\,0.043$) &
\scriptsize($\pm\,0.023$) & \scriptsize($\pm\,0.082$) &
\scriptsize($\pm\,0.003$) & \scriptsize($\pm\,0.043$) &
\scriptsize($\pm\,0.020$) & \scriptsize($\pm\,0.033$) \\
\bottomrule
\end{tabular}%
\label{tab:ablation_style_clustering}
\end{table*}

\begin{table}[!t]
\centering
\renewcommand{\arraystretch}{1.2}
\setlength{\tabcolsep}{5pt}
\caption{Ablation study of gene preprocessing methods.}
\begin{tabular}{@{}c|cc cc@{}}
\toprule
\multirow{2}{*}{\textbf{Preprocessing}} &
\multicolumn{2}{c}{\textbf{TANGLE}} &
\multicolumn{2}{c}{\textbf{MIRROR}} \\
\cmidrule(lr){2-3}\cmidrule(lr){4-5}
& \textbf{Subtyping} & \textbf{Survival} & \textbf{Subtyping} & \textbf{Survival} \\
\midrule
\multirow{2}{*}{PORPOISE} &
0.929 & 0.534 & 0.931 & 0.656 \\
& \multicolumn{1}{c}{\scriptsize($\pm\,0.034$)} &
  \multicolumn{1}{c}{\scriptsize($\pm\,0.099$)} &
  \multicolumn{1}{c}{\scriptsize($\pm\,0.028$)} &
  \multicolumn{1}{c}{\scriptsize($\pm\,0.076$)} \\
\multirow{2}{*}{SurvPath} &
0.931 & 0.535 & 0.938 & 0.628 \\
& \multicolumn{1}{c}{\scriptsize($\pm\,0.014$)} &
  \multicolumn{1}{c}{\scriptsize($\pm\,0.088$)} &
  \multicolumn{1}{c}{\scriptsize($\pm\,0.008$)} &
  \multicolumn{1}{c}{\scriptsize($\pm\,0.079$)} \\
\multirow{2}{*}{MIRROR} &
\textbf{0.952} & \textbf{0.540} & \textbf{0.955} & \textbf{0.665} \\
& \multicolumn{1}{c}{\scriptsize($\pm\,0.022$)} &
  \multicolumn{1}{c}{\scriptsize($\pm\,0.055$)} &
  \multicolumn{1}{c}{\scriptsize($\pm\,0.023$)} &
  \multicolumn{1}{c}{\scriptsize($\pm\,0.082$)} \\
\bottomrule
\end{tabular}%
\label{tab:ablation_gene_preproc}
\end{table}

\begin{figure}[!t]
\centering
\includegraphics[width=0.78\columnwidth]{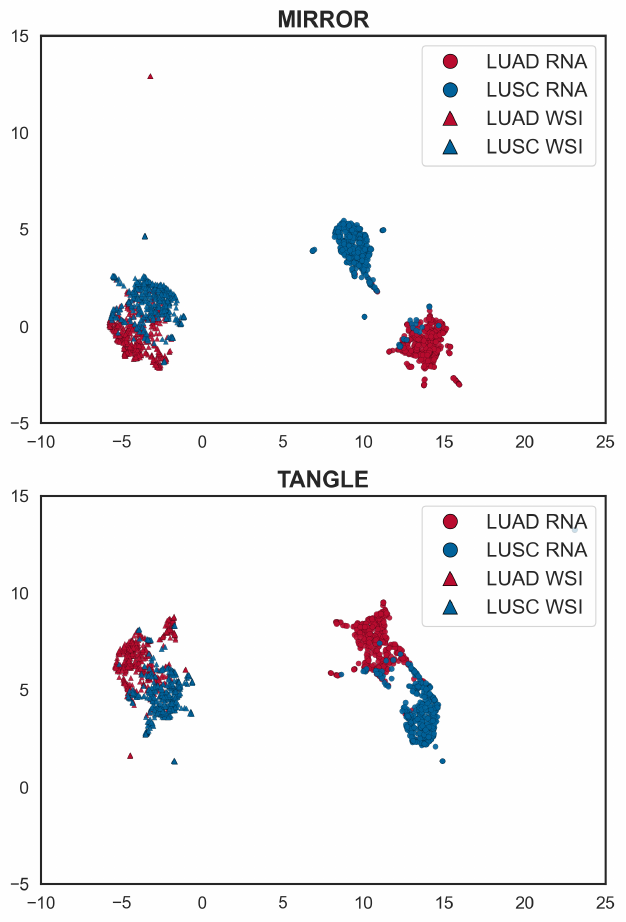}
\caption{\textbf{Visualization of histopathology and transcriptomics features encoded by \proposedmethod\ on the TCGA-NSCLC dataset, compared to those obtained using TANGLE.} Red dots represent samples from TCGA-LUAD, while blue dots represent samples from TCGA-LUSC. Circles denote RNA features, while triangles represent WSI features. \proposedmethod\ clearly yields more distinct and well-aligned feature distributions.}
\label{fig:umap}
\end{figure}

Additionally, as depicted in Figure~\ref{fig:umap}, we visualize features extracted by encoders from TANGLE and \proposedmethod\ on the TCGA-NSCLC dataset. Red and blue dots represent TCGA-LUAD and TCGA-LUSC samples, respectively. Features from \proposedmethod\ exhibit significantly improved class separability, with minimal overlap between subtypes. In contrast, TANGLE's feature space demonstrates considerable inter-class mixing, suggesting weaker discriminative capability. The Euclidean distances between the two modalities are also computed for each method. Notably, the distance for \proposedmethod\ is reduced by 29.02\% compared to TANGLE, indicating a substantially improved alignment between modalities. These findings highlight the advantage of \proposedmethod\ in generating more representative and well-aligned feature distributions, ultimately contributing to superior model performance on downstream tasks.

\begin{figure*}[!t]
\centering
\includegraphics[width=\textwidth]{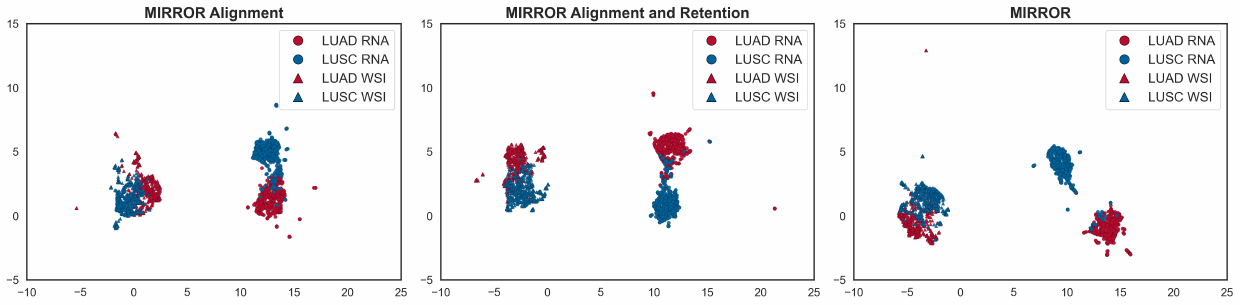}
\caption{\textbf{Visualization of histopathology and transcriptomics features encoded by \proposedmethod\ and its ablated variants on the TCGA-NSCLC dataset.} Red dots represent samples from TCGA-LUAD, while blue dots represent samples from TCGA-LUSC. Circles denote RNA features, while triangles denote WSI features. The comparison clearly illustrates the contribution of each proposed module to the overall feature distributions learned by \proposedmethod.}
\label{fig:umap_ablation}
\end{figure*}

Furthermore, as shown in Figure~\ref{fig:umap_ablation}, we visualize features encoded by \proposedmethod\ and its ablated variants on the TCGA-NSCLC dataset. The variant with the alignment module alone produces compact embeddings and substantially reduces the gap between modalities, thereby achieving strong cross-modal alignment, although subtype boundaries remain indistinct. Incorporating the retention module results in more distinctive clusters and clearer separation between subtypes while preserving the improved alignment. The full model with style clustering further consolidates intra-class cohesion and enhances inter-class separation, consistent with the suppression of nuisance style redundancy and the preservation of disease-relevant information. This progression across variants highlights how the alignment, retention, and style clustering modules collectively contribute to well-aligned and highly discriminative representations.

\section{Discussion}

\subsection{Gene Attribution Analysis}

To gain a deeper understanding of the decision-making process of \proposedmethod\ and to provide potential biological insights into the disease, we conducted gene attribution analysis using Integrated Gradients~\cite{sundararajan2017axiomatic}. Attribution scores were computed over the full set of input genes used during model training.

\begin{figure*}[!t]
\centering
\includegraphics[width=\textwidth]{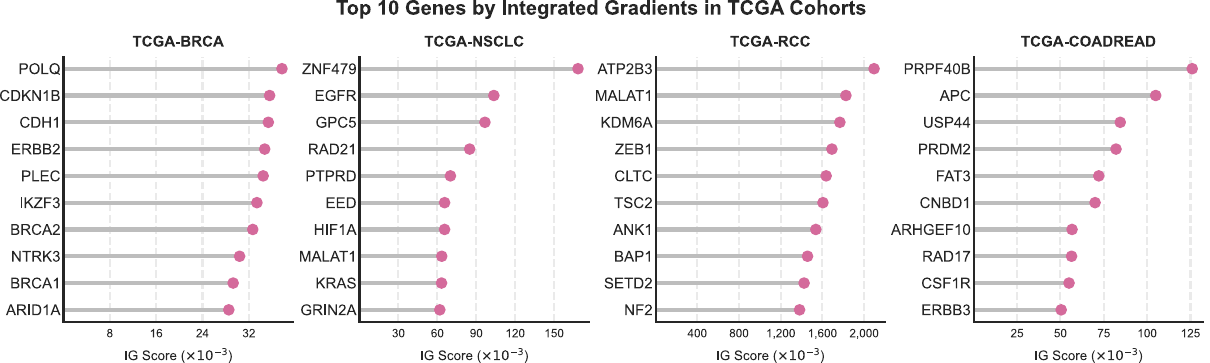}
\caption{\textbf{Top 10 contributing genes identified by \proposedmethod\ across four TCGA cohorts using Integrated Gradients.} Gene attribution scores were computed separately for each cohort, and the top-ranked genes were selected from those annotated in the COSMIC database. These genes highlight the most influential molecular features captured by the model in each cohort.}
\label{fig:gene_ig}
\end{figure*}

To assess biological interpretability, we examined attribution scores for genes annotated in the COSMIC database. As shown in Figure~\ref{fig:gene_ig}, many of the top-ranked genes in each cohort align well with established cancer drivers. In TCGA-BRCA, these include BRCA1/2 and ERBB2~\cite{narod2004brca1,baselga2009novel}, in TCGA-NSCLC, EGFR~\cite{paez2004egfr}, in TCGA-RCC, SETD2 and BAP1~\cite{dalgliesh2010systematic,pena2012bap1}, and in TCGA-COADREAD, APC~\cite{kandoth2013mutational}. When we extended the analysis to the full training gene set, additional high-scoring features were also identified. These include PHF19 and SERPINA5~\cite{brien2012polycomb,asanuma2007protein} in TCGA-BRCA, NFATC4~\cite{yang2025nfatc4} in TCGA-NSCLC, ARPC1B~\cite{tang2023arpc1b} in TCGA-RCC, and PHB1~\cite{alula2021nuclear} in TCGA-COADREAD.

These findings demonstrate the potential of \proposedmethod\ to uncover biologically meaningful patterns from multi-modal data. The use of COSMIC annotations provides a practical framework for interpretability, while future work may further substantiate these findings through external validation or functional assays.

\subsection{Adaptability Across Disease Types}
We evaluate cross-disease adaptation performance to assess how well the learned representations extend beyond the source cohort. For both TANGLE and MIRROR, weights pretrained on TCGA-BRCA with features extracted by Phikon are frozen, and the embedding layer together with a linear classifier is trained on TCGA-NSCLC, TCGA-RCC, and TCGA-COADREAD for subtype classification and survival prediction, using the same 5-fold cross-validation splits utilized in Section~\ref{subsec:downstream_tasks}. The complete results are presented in Tables~\ref{tab:adapt_subtyping} and~\ref{tab:adapt_survival}. On average across the three target cohorts, MIRROR achieves a 5\% improvement in subtype accuracy compared with TANGLE, and increases the C-index by 0.094 for survival prediction. These consistent gains demonstrate strong generalization beyond the source disease and confirm the superiority of MIRROR in out-of-domain transfer.

\begin{table}[!t]
\centering
\renewcommand{\arraystretch}{1.2}
\setlength{\tabcolsep}{5pt}
\caption{Ablation study of loss term weights.}
\begin{tabular}{@{}ccc|cc@{}}
\toprule
\multicolumn{3}{c|}{\textbf{Module Setting}} &
\multicolumn{2}{c}{\textbf{Performance}} \\
\cmidrule(lr){4-5}
\textit{$\lambda_{\alpha}$} & \textit{$\lambda_{\beta}$} & \textit{$\lambda_{\gamma}$} &
\textbf{Subtyping} & \textbf{Survival} \\
\midrule
0.3 & 0.5 & 0.2 & 0.942 {\scriptsize($\pm$ 0.020)} & 0.650 {\scriptsize($\pm$ 0.053)} \\
0.3 & 0.2 & 0.5 & 0.950 {\scriptsize($\pm$ 0.021)} & 0.657 {\scriptsize($\pm$ 0.041)} \\
0.5 & 0.3 & 0.2 & \textbf{0.955} {\scriptsize($\pm$ 0.023)} & \textbf{0.665} {\scriptsize($\pm$ 0.082)} \\
\bottomrule
\end{tabular}%
\label{tab:ablation_loss}
\end{table}

We also observe a limitation on TCGA-COADREAD, where performance reported is substantially lower than that obtained by direct pretraining and adaptation on the TCGA-COADREAD cohort. This likely reflects greater intra-cohort heterogeneity that makes transfer to TCGA-COADREAD more challenging.

To further evaluate generalizability beyond TCGA cohorts, we conduct an external experiment on the Clinical Proteomic Tumor Analysis Consortium Kidney cohort (CPTAC-KIDNEY)~\cite{edwards2015cptac}, which contains 783 slides of clear cell renal cell carcinoma (ccRCC) and 126 slides of non–clear cell renal cell carcinoma (non-ccRCC). The evaluation follows a 5-fold cross-validation protocol consistent with our downstream settings. In contrast to the TCGA-RCC dataset, which is larger and relatively more balanced across subtypes with 519 slides for KIRC, 300 slides for KIRP, and 121 slides for KICH, the CPTAC-KIDNEY is smaller in scale and exhibits a markedly imbalanced subtype distribution, with ccRCC substantially outnumbering non-ccRCC. Despite these challenges, \proposedmethod\ maintains strong transfer performance, outperforming TANGLE on both subtype classification and survival analysis tasks. These results demonstrate \proposedmethod’s robustness to distribution shift across smaller and more heterogeneous datasets.

\subsection{Sensitivity Analysis}
\subsubsection{Sensitivity to Input Perturbations}
To assess the sensitivity of model predictions to small perturbations in histopathology and transcriptomic inputs, we computed two per-sample gradient-based sensitivity metrics on the TCGA-BRCA cohort.

The first sensitivity metric is the $\ell_2$ norm of the input gradient of the logit margin:
\begin{equation}
    s_1(x)\;=\;\bigl\|\nabla_x\!\bigl(f_y(x)-\max_{k\neq y} f_k(x)\bigr)\bigr\|_2,
\end{equation}
where $f_y(x)$ denotes the logit of the true class label $y$ and $\max_{k\neq y} f_k(x)$ is the largest competing logit. This measure captures how rapidly the confidence margin between the correct class and its closest rival changes under infinitesimal input perturbations. A larger $s_1$ value indicates that even small feature perturbations can significantly reduce the margin and move the sample closer to a decision boundary, whereas a smaller value reflects more robust and stable predictions.

Complementary to this margin-based view, the second sensitivity metric considers the overall smoothness of the input--output mapping through the Frobenius norm of the Jacobian:
\begin{equation}
    s_2(x)\;=\;\bigl\|J(x)\bigr\|_F,\qquad J(x)\;=\;\frac{\partial f(x)}{\partial x},
\end{equation}
where $f(x)\in\mathbb{R}^{C}$ denotes the vector of class logits for $C$ classes. The Jacobian $J(x)\in\mathbb{R}^{C\times p}$ collects the partial derivatives of each logit with respect to each input feature, with entry $J_{c,j} = \tfrac{\partial f_c}{\partial x_j}$. Here, $c \in \{1,\dots,C\}$ indexes the class dimension and $j \in \{1,\dots,p\}$ indexes the $p$ input features. The Frobenius norm provides a summary of the aggregate local sensitivity of all logits to small changes in the inputs. Larger values of $s_2$ imply that slight perturbations may lead to substantial variations across the output logits, whereas smaller values indicate a smoother local input--output mapping and more stable behavior~\cite{novak2018sensitivity}.

Based on these two measures, \proposedmethod\ demonstrates substantially lower sensitivity than TANGLE on the TCGA-BRCA cohort, with $s_1=0.768\pm0.045$ compared with $13.101\pm3.029$ and $s_2=0.484\pm0.248$ compared with $8.112\pm4.686$. These results consistently indicate that \proposedmethod\ yields predictions that are less affected by small perturbations in histopathology or transcriptomic inputs and therefore provide greater robustness to potential preprocessing errors or feature noise.

\subsubsection{Sensitivity to Distribution Shifts in Survival Data}
In addition to the local sensitivity analysis, we further examine each model’s sensitivity to distributional factors in the survival analysis task, namely label imbalance and right-censoring, using the TCGA-NSCLC cohort. Following standard practice, survival times are discretized into four bins with sample counts of $190$, $202$, $280$, and $250$, totaling $922$ cases, among which $563$ ($61.1\%$) are censored. To evaluate robustness under these conditions, we compute three censoring-aware metrics on the training folds. The Integrated Brier Score (IBS) reflects the time-integrated prediction error, the mean time-dependent Area Under the Receiver Operating Characteristic Curve (t-AUC) summarizes discriminative ability across time bins, and the mean Brier score represents the average bin-wise squared error. Lower IBS and Brier values together with higher t-AUC values indicate better performance. \proposedmethod\ consistently outperforms TANGLE across all three measures, achieving an IBS of $0.119$ versus $0.217$, a mean t-AUC of $0.940$ versus $0.598$, and a mean Brier score of $0.141$ versus $0.230$. These results demonstrate that \proposedmethod\ attains stronger discriminative power and lower prediction error despite substantial censoring and label imbalance in the cohort.

\subsection{Computational Complexity}
We further analyzed the computational cost of \proposedmethod\ to assess the trade-offs introduced by the proposed architecture and modules. The number of parameters, floating-point operations (FLOPs) per sample, runtime throughput, and memory usage were measured and reported to provide a comprehensive evaluation.

When applied to downstream tasks, where a single linear classifier is attached to the backbone, \proposedmethod\ comprises $33.10$M parameters and requires $90.44$G FLOPs per forward pass. During pretraining, the complexity increases to $48.97$M parameters and $139.14$G FLOPs, reflecting the additional overhead of these components. In comparison, TANGLE remains considerably lighter in both pretraining and downstream evaluation, with $11.41$M parameters and $9.69$G FLOPs. For runtime efficiency, we further measured throughput and peak GPU memory usage on our experimental environment. At inference with a batch size of 16, \proposedmethod\ processes $227.88$ samples/s, whereas TANGLE reaches $2{,}593.77$ samples/s. During pretraining, also with a batch size of 16, \proposedmethod\ attains $61.58$ samples/s per optimization step, while TANGLE achieves $799.19$ samples/s. The peak GPU memory consumption during training is $10.30$\,GB for the \proposedmethod\,, $19.24$\,GB for the pretraining model, and $3.17$\,GB for TANGLE.

These results indicate that \proposedmethod\ improves robustness and multimodal integration at the expense of a larger memory footprint and greater computational demand. This trade-off highlights a limitation of the current design and points to opportunities for future work.

\subsection{Robustness to Domain Shift}
Our experiments are conducted on four TCGA cohorts, namely TCGA-BRCA, TCGA-NSCLC, TCGA-RCC, and TCGA-COADREAD, which collectively encompass three principal sources of domain variability: institutional variation, scanner heterogeneity, and artifacts in formalin-fixed paraffin-embedded (FFPE) slides. This setup inherently evaluates the robustness of \proposedmethod\ under realistic and diverse domain shifts encountered in clinical applications.

To capture institutional variability, TCGA aggregates histopathology slides from a large network of contributing sites~\cite{dehkharghanian2023biased}. This multi-center composition introduces substantial heterogeneity in staining protocols, fixation methods, and tissue processing workflows. Scanner hardware also varies across sites, with both Aperio and Hamamatsu systems used in the digitization process~\cite{vaidya2025molecular}, leading to differences in image resolution, color reproduction, and compression artifacts. FFPE slide preparation further introduces artifacts such as tissue folds, air bubbles, slide-edge cracks, out-of-focus regions, and pen markings. Recent quality control audits~\cite{weng2024grandqc, janowczyk2019histoqc, kassab2024ffpe++} provide quantitative evidence of the prevalence and diversity of these artifacts in TCGA cohorts.

Our 5-fold cross-validation strategy ensures that each fold includes diverse institutions, scanner types, and slide preparation modes. As a result, the reported performance reflects \proposedmethod's generalization under realistic, multi-center deployment conditions.

\begin{table}[!t]
\centering
\renewcommand{\arraystretch}{1.2}
\setlength{\tabcolsep}{2pt}
\caption{Adaptation performance for cancer subtype classification using TCGA-BRCA pretrained weights.}
\resizebox{\columnwidth}{!}{%
\begin{tabular}{@{}c|cc cc cc cc@{}}
\toprule
\multirow{2}{*}{\textbf{Method}} &
\multicolumn{2}{c}{\shortstack{\textbf{TCGA-}\\\textbf{NSCLC}}} &
\multicolumn{2}{c}{\shortstack{\textbf{TCGA-}\\\textbf{RCC}}} &
\multicolumn{2}{c}{\shortstack{\textbf{TCGA-}\\\textbf{COADREAD}}} &
\multicolumn{2}{c}{\shortstack{\textbf{CPTAC-}\\\textbf{KIDNEY}}} \\
\cmidrule(lr){2-3}\cmidrule(lr){4-5}\cmidrule(lr){6-7}\cmidrule(lr){8-9}
 & \textbf{Acc} & \textbf{F1} & \textbf{Acc} & \textbf{F1} & \textbf{Acc} & \textbf{F1} & \textbf{Acc} & \textbf{F1} \\
\midrule
\multirow{2}{*}{TANGLE} &
0.947 & 0.987 & 0.956 & 0.908 & 0.744 & 0.448 & 0.934 & 0.827 \\
& \multicolumn{1}{c}{\scriptsize($\pm\,0.044$)} &
  \multicolumn{1}{c}{\scriptsize($\pm\,0.110$)} &
  \multicolumn{1}{c}{\scriptsize($\pm\,0.004$)} &
  \multicolumn{1}{c}{\scriptsize($\pm\,0.148$)} &
  \multicolumn{1}{c}{\scriptsize($\pm\,0.175$)} &
  \multicolumn{1}{c}{\scriptsize($\pm\,0.020$)} &
  \multicolumn{1}{c}{\scriptsize($\pm\,0.005$)} &
  \multicolumn{1}{c}{\scriptsize($\pm\,0.011$)} \\
\multirow{2}{*}{MIRROR} &
\textbf{0.993} & \textbf{0.993} & \textbf{0.983} & \textbf{0.914} & \textbf{0.822} & \textbf{0.523} & \textbf{0.986} & \textbf{0.964} \\
& \multicolumn{1}{c}{\scriptsize($\pm\,0.006$)} &
  \multicolumn{1}{c}{\scriptsize($\pm\,0.006$)} &
  \multicolumn{1}{c}{\scriptsize($\pm\,0.012$)} &
  \multicolumn{1}{c}{\scriptsize($\pm\,0.145$)} &
  \multicolumn{1}{c}{\scriptsize($\pm\,0.050$)} &
  \multicolumn{1}{c}{\scriptsize($\pm\,0.075$)} &
  \multicolumn{1}{c}{\scriptsize($\pm\,0.001$)} &
  \multicolumn{1}{c}{\scriptsize($\pm\,0.002$)} \\
\bottomrule
\end{tabular}}
\label{tab:adapt_subtyping}
\end{table}

\subsection{Limitations and Future Work}
In this section, we reflect on the limitations of the present study and outline directions that would strengthen both performance and clinical readiness.

The current evaluation is confined to TCGA, which, although extensive and spanning multiple cancer types, does not fully capture the diversity of clinical workflows and patient populations. Expanding to external datasets will provide a more comprehensive assessment of robustness under real-world variability and yield more abundant training and evaluation samples. At the same time, pretraining is currently restricted to intra-cohort data. Extending pretraining across all TCGA cohorts and, where feasible, beyond TCGA would enable the development of a stronger and more reliable multimodal foundation model. Another limitation arises from transcriptomic inputs, which differ in length across cohorts because the preprocessing pipeline selects different gene subsets for each cohort. Establishing a harmonized gene panel across TCGA, together with the integration of pathway-level and additional omics variables using consistent preprocessing, would facilitate the training of larger and more coherent foundation models.

Beyond data-related constraints, computational efficiency also poses a challenge. The heavier backbone and modules lead to a higher parameter count and increased FLOPs compared with simpler baselines. To broaden deployability in settings with limited latency and memory budgets, future work will investigate parameter-efficient fine-tuning, knowledge distillation, and quantization in order to produce a family of models that vary in size and computational cost. In addition, practical deployments often encounter incomplete or noisy inputs, such as missing RNA profiles or variable patch-level feature quality. Addressing these issues will require modality-robust training strategies and uncertainty-aware mechanisms that can ensure reliable predictions even under imperfect input conditions.

Taken together, these directions address the principal limitations related to data breadth, modality robustness, input consistency, and computational efficiency. Advancing along these lines is expected to improve overall performance and generalization under domain shift, enhance stability and interpretability, and ultimately move \proposedmethod\ closer to serving as a clinically viable multimodal foundation model ready for deployment across diverse healthcare environments.

\begin{table}[!t]
\centering
\renewcommand{\arraystretch}{1.2}
\setlength{\tabcolsep}{2pt}
\caption{Adaptation performance for cancer survival analysis using TCGA-BRCA pretrained weights.}
\resizebox{\columnwidth}{!}{%
\begin{tabular}{@{}c|cccc@{}}
\toprule
\multicolumn{1}{c|}{\textbf{Method}} &
\shortstack{\textbf{TCGA-}\\\textbf{NSCLC}} &
\shortstack{\textbf{TCGA-}\\\textbf{RCC}} &
\shortstack{\textbf{TCGA-}\\\textbf{COADREAD}} &
\shortstack{\textbf{CPTAC-}\\\textbf{KIDNEY}} \\
\midrule
TANGLE &
0.536 {\scriptsize($\pm\,0.025$)} &
0.744 {\scriptsize($\pm\,0.130$)} &
0.511 {\scriptsize($\pm\,0.016$)} &
0.614 {\scriptsize($\pm\,0.087$)}\\
MIRROR &
\textbf{0.602} {\scriptsize($\pm\,0.044$)} &
\textbf{0.778} {\scriptsize($\pm\,0.055$)} &
\textbf{0.693} {\scriptsize($\pm\,0.045$)} &
\textbf{0.816} {\scriptsize($\pm\,0.060$)} \\
\bottomrule
\end{tabular}}
\label{tab:adapt_survival}
\end{table}

\section{Conclusion}
In this paper, we introduced \proposedmethod, a novel multi-modal pathological SSL framework designed for joint pretraining of histopathology and transcriptomics data. Our approach is tailored to align modality-shared information while retaining modality-specific unique features. Additionally, it employs a style clustering module to reduce redundancy and preserve disease-relevant representations. Furthermore, we proposed a novel transcriptomics data preprocessing pipeline to efficiently identify disease-related genes, resulting in refined, disease-focused transcriptomics datasets. The proposed method was trained and evaluated on four distinct cohorts from the TCGA dataset, demonstrating superior performance and underscoring the efficacy of our design. Future work will focus on expanding evaluations to external datasets and additional disease cohorts, extending pretraining across and beyond TCGA, and establishing harmonized molecular inputs to support larger foundation models. Efforts will also prioritize reducing computational overhead through parameter-efficient fine-tuning, knowledge distillation, and quantization, as well as developing strategies that enhance robustness to missing or noisy modalities.

\appendices

\section{Proof of Uniqueness for the Global Minimum of the Style Loss}
\label{app:style-proof}

\begin{proposition}\label{prop:style-min}
Let the two modality-specific posteriors be defined as:
\begin{equation}
    \mathcal N(\mu_S, \sigma_S^2 I),
    \quad
    \mathcal N(\mu_T, \sigma_T^2 I),
\end{equation}
with diagonal covariance in $\mathbb R^{D}$. The style loss:
\begin{equation}
\begin{split}
    L_{\text{style}} 
    &=\; \mathrm{KL}\bigl(\mathcal{N}(\mu_S, \sigma_S^2 I) \,\big\Vert\, \mathcal{N}(0,I)\bigr) \\
        &+\, \mathrm{KL}\bigl(\mathcal{N}(\mu_T, \sigma_T^2 I) \,\big\Vert\, \mathcal{N}(0,I)\bigr),
\end{split}
\tag{15}
\end{equation}
has a unique global minimum:
\begin{equation}
    L_{\text{style}}^{\min} = 0,
\end{equation}
attained if and only if:
\begin{equation}
    \mu_S=\mu_T=0,
    \quad
    \sigma_S=\sigma_T=1 .
\end{equation}
\end{proposition}

\begin{proof}
Consider a single latent coordinate $(\mu,\sigma)$, the KL divergence between $\mathcal{N}(\mu, \sigma^2)$ and the standard normal distribution $\mathcal{N}(0,1)$ is given by:
\begin{equation}
    \mathrm{KL}\bigl(\mathcal N(\mu,\sigma^{2})\Vert\mathcal N(0,1)\bigr)
   =f(\mu,\sigma)
   :=\tfrac12\bigl(\mu^{2}+\sigma^{2}-1-\log\sigma^{2}\bigr).
\end{equation}

The first–order derivatives are:
\begin{equation}
    \partial_{\mu} f = \mu,
    \quad
    \partial_{\sigma} f = \sigma - \sigma^{-1},
\end{equation}
which both vanish if and only if $(\mu,\sigma)=(0,1)$. Furthermore, the Hessian is given by:
\begin{equation}
    \nabla^{2}f(\mu,\sigma)=
    \begin{pmatrix}
    1 & 0\\[2pt]
    0 & 1+\sigma^{-2}
    \end{pmatrix}\succ 0,
\end{equation}
indicating that $f$ is strictly convex. Therefore, $(0,1)$ is the unique global minimizer of $f$, with $f(0,1)=0$.

Since the KL divergence between diagonal Gaussians decomposes as a sum over coordinates, the total style loss becomes:
\begin{equation}
    L_{\text{style}}
   =\tfrac12\sum_{i=1}^{D}
       \bigl(\mu_{S,i}^{2}+\sigma_{S,i}^{2}-1-\log\sigma_{S,i}^{2}
            +\mu_{T,i}^{2}+\sigma_{T,i}^{2}-1-\log\sigma_{T,i}^{2}\bigr).
\end{equation}
Each summand is strictly convex and minimized uniquely at
$\mu_{S,i} = \mu_{T,i} = 0$ and $\sigma_{S,i} = \sigma_{T,i} = 1$. Thus, the global minimum $L_{\text{style}} = 0$ is achieved if and only if:
\begin{equation}
    \mu_S = \mu_T = 0,
    \quad
    \sigma_S = \sigma_T = 1.
\end{equation}
\end{proof}

\section{Proof for Minimizing the Clustering Loss Increases Mutual Information}
\label{app:cluster-proof}

\begin{proposition}\label{prop:cluster-mi}
Let $\mathbf S_{\text{cluster}},\mathbf T_{\text{cluster}}\in\Delta^{K-1}$ denote the soft prototype assignments for the two modalities, and let $z_S,z_T\in\mathbb R^{D}$ be the corresponding style representations.  Define the clustering loss as:
\begin{equation}
\begin{aligned}
L_{\text{cluster}}
   &= \mathrm{KL}\!\bigl(\mathbf{S}_{\text{cluster}}
                         \,\|\, 
                         \mathbf{T}_{\text{cluster}}\bigr)
      + \mathrm{KL}\!\bigl(\mathbf{T}_{\text{cluster}}
                           \,\|\, 
                           \mathbf{S}_{\text{cluster}}\bigr)\\
   &= 2\,\mathrm{JSD}\!\bigl(\mathbf{S}_{\text{cluster}},
                             \mathbf{T}_{\text{cluster}}\bigr).
\end{aligned}
\tag{19}
\end{equation}
and define the function:
\begin{equation}
g(L):=\sqrt{\frac{L}{2}}
       \Bigl[\log\!\frac{K-1}{\sqrt{L/2}}+1\Bigr],
       \quad 0<L\le 2\log 2.    
\end{equation}
Then, the mutual information between $z_S$ and $z_T$ satisfies:
\begin{equation}
-\,I(z_S; z_T) \;\le\; g\bigl(L_{\text{cluster}}\bigr),
\end{equation}
where $g(\cdot)$ is strictly increasing. Consequently, minimizing the clustering loss $L_{\text{cluster}}$ strictly increases the mutual information shared by the two modalities.
\end{proposition}

\begin{proof}
Since $L_{\mathrm{cluster}}\ge \mathrm{KL}(\mathbf S_{\text{cluster}}\| \mathbf T_{\text{cluster}})$, Pinsker’s inequality~\cite{pinsker1960information} gives:
\begin{equation}
\label{eq:pinsker}
\|\mathbf S_{\text{cluster}}-\mathbf T_{\text{cluster}}\|_1
   \le\sqrt{2L_{\mathrm{cluster}}}.
\end{equation}

To relate this bound to a mismatch probability, consider the two categorical distributions $\mathbf S_{\text{cluster}}=(s_1,\dots,s_K)$ and $\mathbf T_{\text{cluster}}=(t_1,\dots,t_K)$. For each index $k$, define:
\begin{equation}
\begin{split}
    r_k &:= \min(s_k, t_k), \\
    a_k &:= \max(s_k - t_k, 0), \\
    b_k &:= \max(t_k - s_k, 0),
\end{split}
\end{equation}
so that $r_k$ is the shared mass, and $a_k$ and $b_k$ are the excess contributions from Slide and Transcriptomics modalities, respectively. Let $r := \sum_k r_k$, then $\sum_k a_k=\sum_k b_k = 1 - r$ and $\lVert\mathbf S_{\text{cluster}}-\mathbf T_{\text{cluster}}\rVert_1=2(1-r)$.

Now construct a coupling $C^\ast=(c_{kl}^\ast)$ as follows: assign the shared mass to the diagonal, $c_{kk}^\ast=r_k$. Since the total surplus is equal on both sides, define a bijection $\pi:\{k:a_k>0\}\to\{k:b_k>0\}$, and let:
\begin{equation}
    c^*_{k, \pi(k)} = \tfrac{1}{2} a_k,
    \quad
    c^*_{\pi(k), k} = \tfrac{1}{2} b_{\pi(k)}.
\end{equation}
Set all remaining entries to zero. By construction, $C^\ast$ is a valid joint distribution with marginals $\mathbf{S}_{\mathrm{cluster}}$ and $\mathbf{T}_{\mathrm{cluster}}$.

Under this coupling, disagreement occurs only in the surplus mass. Hence, the probability of mismatch is:
\begin{equation}
\label{eq:optimal_coupling}
\begin{split}
  p_e&:=\Pr_{C^\ast}[K_S\neq K_T] \\
      &=1-r \\
      &=\tfrac12\lVert\mathbf S_{\text{cluster}}
                      -\mathbf T_{\text{cluster}}\rVert_1 \\
      &=d_{\mathrm{TV}}\bigl(\mathbf S_{\text{cluster}},
                            \mathbf T_{\text{cluster}}\bigr).
\end{split}
\end{equation}
Combining Eq.~\eqref{eq:optimal_coupling} with Eq.~\eqref{eq:pinsker}, we have:
\begin{equation}
\label{eq:pinsker_optimal_coupling}
p_e\;\le\;\sqrt{\tfrac{L_{\mathrm{cluster}}}{2}}.
\end{equation}

Applying Fano’s inequality~\cite{fano1961transmission}, we have:
\begin{equation}
\begin{split}
    H(K_S\!\mid K_T)
  &\le h_2(p_e)+p_e\log(K-1) \\
  &\le p_e\Bigl[\log\tfrac{K-1}{p_e}+1\Bigr],
\end{split}
\end{equation}
where $h_2(\cdot)$ is the binary entropy function. Substituting the bound Eq.~\eqref{eq:pinsker_optimal_coupling}, we obtain:
\begin{equation}
    H(K_S\!\mid K_T) \le g\!\bigl(L_{\mathrm{cluster}}\bigr).
\end{equation}
Since $I(K_S;K_T)=H(K_S)-H(K_S\!\mid K_T)$ and $H(K_S)\ge 0$, we have:
\begin{equation}
    -\;I(K_S;K_T)\;\le\;g\!\bigl(L_{\mathrm{cluster}}\bigr).
\end{equation}

Note that $L_{\mathrm{cluster}} = 2\,\mathrm{JSD}(\mathbf{S}_{\mathrm{cluster}}, \mathbf{T}_{\mathrm{cluster}})$, and since $\mathrm{JSD}(P, Q) \le \log 2$ for any pair of distributions $P, Q$, it follows that:
\begin{equation}
    0 < L_{\mathrm{cluster}} \le 2 \log 2.
\end{equation}
This justifies the domain restriction in the definition of $g(L)$.
Now observe that:
\begin{equation}
    g'(L) = \frac{1}{2\sqrt{2L}} \left[\log(K - 1) + \tfrac{1}{2} \log 2 - \tfrac{1}{2} \log L \right] > 0
\end{equation}
for $0 < L \le 2 \log 2$, so $g(\cdot)$ is strictly increasing.

Because the prototype assignments $K_S$ and $K_T$ are deterministic functions of the latents $z_S$ and $z_T$, the data processing inequality~\cite{cover2006elements} implies:
\begin{equation}
    I(K_S;K_T) \le I(z_S;z_T).
\end{equation}
Taking negatives and combining with the previous bound yields:
\begin{equation}
    -\;I(z_S;z_T)
      \;\le\;
    -\;I(K_S;K_T)
      \;\le\;
    g\!\bigl(L_{\mathrm{cluster}}\bigr).
\end{equation}

Since $g(\cdot)$ is strictly increasing, reducing $L_{\text{cluster}}$ strictly reduces the upper bound, thereby increasing the mutual information $I(z_S; z_T)$.
\end{proof}

\bibliographystyle{IEEEtran}
\bibliography{ieee}

\begin{thebibliography}{10}
\providecommand{\url}[1]{#1}
\csname url@samestyle\endcsname
\providecommand{\newblock}{\relax}
\providecommand{\bibinfo}[2]{#2}
\providecommand{\BIBentrySTDinterwordspacing}{\spaceskip=0pt\relax}
\providecommand{\BIBentryALTinterwordstretchfactor}{4}
\providecommand{\BIBentryALTinterwordspacing}{\spaceskip=\fontdimen2\font plus
\BIBentryALTinterwordstretchfactor\fontdimen3\font minus \fontdimen4\font\relax}
\providecommand{\BIBforeignlanguage}[2]{{%
\expandafter\ifx\csname l@#1\endcsname\relax
\typeout{** WARNING: IEEEtran.bst: No hyphenation pattern has been}%
\typeout{** loaded for the language `#1'. Using the pattern for}%
\typeout{** the default language instead.}%
\else
\language=\csname l@#1\endcsname
\fi
#2}}
\providecommand{\BIBdecl}{\relax}
\BIBdecl

\bibitem{chang2007evidence}
A.~E. Chang \emph{et~al.}, \emph{An evidence-based approach}.\hskip 1em plus 0.5em minus 0.4em\relax New York, NY, USA: Springer, 2007, 34.

\bibitem{7886294}
Y.~Song, Q.~Li, H.~Huang, D.~Feng, M.~Chen, and W.~Cai, ``Low dimensional representation of fisher vectors for microscopy image classification,'' \emph{IEEE Trans. Med. Imaging}, 2017, 36(8):1636-49.

\bibitem{chen2022scaling}
R.~J. Chen \emph{et~al.}, ``Scaling vision transformers to gigapixel images via hierarchical self-supervised learning,'' in \emph{CVPR}, 2022, :16144-55.

\bibitem{chen2024towards}
R.~J. Chen \emph{et~al.}, ``Towards a general-purpose foundation model for computational pathology,'' \emph{Nature Medicine}, 2024, 30(3):850-62.

\bibitem{xiang2022dsnet}
T.~Xiang \emph{et~al.}, ``Dsnet: A dual-stream framework for weakly-supervised gigapixel pathology image analysis,'' \emph{IEEE Trans. Med. Imaging}, 2022, 41(8):2180-90.

\bibitem{li2023single}
Z.~Li, Y.~Jiang, L.~Liu, Y.~Xia, and R.~Li, ``Single-cell spatial analysis of histopathology images for survival prediction via graph attention network,'' in \emph{International Workshop on Applications of Medical AI}, 2023, :114-24.

\bibitem{saiki1985enzymatic}
R.~K. Saiki \emph{et~al.}, ``Enzymatic amplification of $\beta$-globin genomic sequences and restriction site analysis for diagnosis of sickle cell anemia,'' \emph{Science}, 1985, 230(4732):1350-54.

\bibitem{radford2021learning}
A.~Radford \emph{et~al.}, ``Learning transferable visual models from natural language supervision,'' in \emph{ICML}, 2021, 8748-63.

\bibitem{wang2022medclip}
Z.~Wang, Z.~Wu, D.~Agarwal, and J.~Sun, ``Medclip: Contrastive learning from unpaired medical images and text,'' \emph{arXiv preprint arXiv:2210.10163}, 2022.

\bibitem{xu2024multimodal}
Y.~Xu \emph{et~al.}, ``A multimodal knowledge-enhanced whole-slide pathology foundation model,'' \emph{arXiv preprint arXiv:2407.15362}, 2024.

\bibitem{boehm2022harnessing}
K.~M. Boehm, P.~Khosravi, R.~Vanguri, J.~Gao, and S.~P. Shah, ``Harnessing multimodal data integration to advance precision oncology,'' \emph{Nat. Rev. Cancer}, 2022, 22(2):114-26.

\bibitem{jaume2024transcriptomics}
G.~Jaume \emph{et~al.}, ``Transcriptomics-guided slide representation learning in computational pathology,'' in \emph{CVPR}, 2024, :9632-44.

\bibitem{zhang2024prototypical}
Y.~Zhang, Y.~Xu, J.~Chen, F.~Xie, and H.~Chen, ``Prototypical information bottlenecking and disentangling for multimodal cancer survival prediction,'' \emph{arXiv preprint arXiv:2401.01646}, 2024.

\bibitem{jaume2024modeling}
G.~Jaume, A.~Vaidya, R.~J. Chen, D.~F. Williamson, P.~P. Liang, and F.~Mahmood, ``Modeling dense multimodal interactions between biological pathways and histology for survival prediction,'' in \emph{CVPR}, 2024, :11579-590.

\bibitem{vaidya2025molecular}
A.~Vaidya \emph{et~al.}, ``Molecular-driven foundation model for oncologic pathology,'' \emph{arXiv preprint arXiv:2501.16652}, 2025.

\bibitem{vaswani2017attention}
A.~Vaswani \emph{et~al.}, ``Attention is all you need,'' \emph{NeurIPS}, 2017, 30.

\bibitem{tomczak2015review}
K.~Tomczak, P.~Czerwi{\'n}ska, and M.~Wiznerowicz, ``Review the cancer genome atlas (tcga): an immeasurable source of knowledge,'' \emph{Contemporary Oncology/Wsp{\'o}{\l}czesna Onkologia}, 2015, 2015(1):68-77.

\bibitem{chen2020simple}
T.~Chen, S.~Kornblith, M.~Norouzi, and G.~Hinton, ``A simple framework for contrastive learning of visual representations,'' in \emph{ICML}, 2020, :1597-1607.

\bibitem{he2020momentum}
K.~He, H.~Fan, Y.~Wu, S.~Xie, and R.~Girshick, ``Momentum contrast for unsupervised visual representation learning,'' in \emph{CVPR}, 2020, :9729-38.

\bibitem{bao2021beit}
H.~Bao, L.~Dong, S.~Piao, and F.~Wei, ``Beit: Bert pre-training of image transformers,'' \emph{arXiv preprint arXiv:2106.08254}, 2021.

\bibitem{he2022masked}
K.~He, X.~Chen, S.~Xie, Y.~Li, P.~Doll{\'a}r, and R.~Girshick, ``Masked autoencoders are scalable vision learners,'' in \emph{CVPR}, 2022, :16000-9.

\bibitem{zhou2021ibot}
J.~Zhou \emph{et~al.}, ``ibot: Image bert pre-training with online tokenizer,'' \emph{arXiv preprint arXiv:2111.07832}, 2021.

\bibitem{oquab2023dinov2}
M.~Oquab \emph{et~al.}, ``Dinov2: Learning robust visual features without supervision,'' \emph{arXiv preprint arXiv:2304.07193}, 2023.

\bibitem{caron2020unsupervised}
M.~Caron, I.~Misra, J.~Mairal, P.~Goyal, P.~Bojanowski, and A.~Joulin, ``Unsupervised learning of visual features by contrasting cluster assignments,'' \emph{NeurIPS}, 2020, 33:9912-24.

\bibitem{jia2021scaling}
C.~Jia \emph{et~al.}, ``Scaling up visual and vision-language representation learning with noisy text supervision,'' in \emph{ICML}, 2021, :4904-16.

\bibitem{alayrac2022flamingo}
J.-B. Alayrac \emph{et~al.}, ``Flamingo: a visual language model for few-shot learning,'' \emph{NeurIPS}, 2022, 35:23716-36.

\bibitem{liu2020pdam}
D.~Liu \emph{et~al.}, ``Pdam: A panoptic-level feature alignment framework for unsupervised domain adaptive instance segmentation in microscopy images,'' \emph{IEEE Trans. Med. Imaging}, 2020, 40(1):154-65.

\bibitem{lin2022label}
Y.~Lin \emph{et~al.}, ``Label propagation for annotation-efficient nuclei segmentation from pathology images,'' \emph{arXiv preprint arXiv:2202.08195}, 2022.

\bibitem{li2023task}
H.~Li \emph{et~al.}, ``Task-specific fine-tuning via variational information bottleneck for weakly-supervised pathology whole slide image classification,'' in \emph{CVPR}, 2023, :7454-63.

\bibitem{fan2024revisiting}
J.~Fan \emph{et~al.}, ``Revisiting adaptive cellular recognition under domain shifts: A contextual correspondence view,'' in \emph{ECCV}, 2024, :275-92.

\bibitem{fan2024seeing}
J.~Fan, D.~Liu, H.~Chang, H.~Huang, M.~Chen, and W.~Cai, ``Seeing unseen: Discover novel biomedical concepts via geometry-constrained probabilistic modeling,'' in \emph{CVPR}, 2024, :11524-34.

\bibitem{fan2025structuring}
J.~Fan, D.~Liu, H.~Chang, H.~Huang, M.~Chen, and W.~Cai, ``On structuring hyperspherical manifold for probing novel biomedical entities,'' \emph{IEEE Trans. Pattern Anal. Mach. Intell.}, 2025.

\bibitem{wang2022transformer}
X.~Wang \emph{et~al.}, ``Transformer-based unsupervised contrastive learning for histopathological image classification,'' \emph{Medical Image Analysis}, Oct. 2022, 81:102559.

\bibitem{ciga2022self}
O.~Ciga, T.~Xu, and A.~L. Martel, ``Self supervised contrastive learning for digital histopathology,'' \emph{Machine Learning with Applications}, Mar. 2022, 7:100198.

\bibitem{ahmed2024pathalign}
F.~Ahmed \emph{et~al.}, ``Pathalign: A vision-language model for whole slide images in histopathology,'' \emph{arXiv preprint arXiv:2406.19578}, 2024.

\bibitem{sun2024cpath}
Y.~Sun \emph{et~al.}, ``Cpath-omni: A unified multimodal foundation model for patch and whole slide image analysis in computational pathology,'' \emph{arXiv preprint arXiv:2412.12077}, 2024.

\bibitem{yang2023mrm}
Q.~Yang, W.~Li, B.~Li, and Y.~Yuan, ``Mrm: Masked relation modeling for medical image pre-training with genetics,'' in \emph{ICCV}, 2023, :21452-62.

\bibitem{lu2023multi}
M.~Lu, T.~Wang, and Y.~Xia, ``Multi-modal pathological pre-training via masked autoencoders for breast cancer diagnosis,'' in \emph{MICCAI}, 2023, :457-66.

\bibitem{xu2024whole}
H.~Xu \emph{et~al.}, ``A whole-slide foundation model for digital pathology from real-world data,'' \emph{Nature}, 2024, :1-8.

\bibitem{vorontsov2023virchow}
E.~Vorontsov \emph{et~al.}, ``Virchow: a million-slide digital pathology foundation model,'' \emph{arXiv preprint arXiv:2309.07778}, 2023.

\bibitem{koohbanani2021self}
N.~A. Koohbanani, B.~Unnikrishnan, S.~A. Khurram, P.~Krishnaswamy, and N.~Rajpoot, ``Self-path: Self-supervision for classification of pathology images with limited annotations,'' \emph{IEEE Trans. Med. Imaging}, 2021, 40(10):2845-56.

\bibitem{singhal2023large}
K.~Singhal \emph{et~al.}, ``Large language models encode clinical knowledge,'' \emph{Nature}, 2023, 620(7972):172-80.

\bibitem{lu2024visual}
M.~Y. Lu \emph{et~al.}, ``A visual-language foundation model for computational pathology,'' \emph{Nature Medicine}, 2024, 30(3):863-74.

\bibitem{li2024llava}
C.~Li \emph{et~al.}, ``Llava-med: Training a large language-and-vision assistant for biomedicine in one day,'' \emph{NeurIPS}, pp. 28\,541--28\,564, 2024, 36.

\bibitem{jaume2024multistain}
G.~Jaume \emph{et~al.}, ``Multistain pretraining for slide representation learning in pathology,'' \emph{arXiv preprint arXiv:2408.02859}, 2024.

\bibitem{yu2022coca}
J.~Yu, Z.~Wang, V.~Vasudevan, L.~Yeung, M.~Seyedhosseini, and Y.~Wu, ``Coca: Contrastive captioners are image-text foundation models,'' \emph{arXiv preprint arXiv:2205.01917}, 2022.

\bibitem{guo2024histgen}
Z.~Guo, J.~Ma, Y.~Xu, Y.~Wang, L.~Wang, and H.~Chen, ``Histgen: Histopathology report generation via local-global feature encoding and cross-modal context interaction,'' in \emph{MICCAI}, 2024, :189-99.

\bibitem{chen2020pathomic}
R.~J. Chen \emph{et~al.}, ``Pathomic fusion: an integrated framework for fusing histopathology and genomic features for cancer diagnosis and prognosis,'' \emph{IEEE Trans. Med. Imaging}, 2020, 41(41):757-70.

\bibitem{zheng2024graph}
Y.~Zheng \emph{et~al.}, ``Graph attention-based fusion of pathology images and gene expression for prediction of cancer survival,'' \emph{IEEE Trans. Med. Imaging}, 2024, 43(9):3085-97.

\bibitem{wang2025histo}
Z.~Wang, Y.~Zhang, Y.~Xu, S.~Imoto, H.~Chen, and J.~Song, ``Histo-genomic knowledge association for cancer prognosis from histopathology whole slide images,'' \emph{IEEE Trans. Med. Imaging}, 2025, :1.

\bibitem{xing2024comprehensive}
X.~Xing, M.~Zhu, Z.~Chen, and Y.~Yuan, ``Comprehensive learning and adaptive teaching: Distilling multi-modal knowledge for pathological glioma grading,'' \emph{Medical Image Analysis}, vol.~91, p. 102990, Jan. 2024.

\bibitem{shao2021transmil}
Z.~Shao \emph{et~al.}, ``Transmil: Transformer based correlated multiple instance learning for whole slide image classification,'' \emph{NeurIPS}, 2021, 34:2136-47.

\bibitem{guyon2002gene}
I.~Guyon, J.~Weston, S.~Barnhill, and V.~Vapnik, ``Gene selection for cancer classification using support vector machines,'' \emph{Machine learning}, 2002, 46:389-422.

\bibitem{sondka2024cosmic}
Z.~Sondka \emph{et~al.}, ``Cosmic: a curated database of somatic variants and clinical data for cancer,'' \emph{Nucleic Acids Research}, 2024, 52(D1):D1210-7.

\bibitem{oord2018representation}
A.~v.~d. Oord, Y.~Li, and O.~Vinyals, ``Representation learning with contrastive predictive coding,'' \emph{arXiv preprint arXiv:1807.03748}, 2018.

\bibitem{kingma2013auto}
D.~P. Kingma and M.~Welling, ``Auto-encoding variational bayes,'' \emph{arXiv preprint arXiv:1312.6114}, 2013.

\bibitem{kullback1951information}
S.~Kullback and R.~A. Leibler, ``On information and sufficiency,'' \emph{The annals of mathematical statistics}, 1951, 22(1):79-86.

\bibitem{tishby2000information}
N.~Tishby, F.~C. Pereira, and W.~Bialek, ``The information bottleneck method,'' \emph{arXiv preprint physics/0004057}, 2000.

\bibitem{alemi2016deep}
A.~A. Alemi, I.~Fischer, J.~V. Dillon, and K.~Murphy, ``Deep variational information bottleneck,'' \emph{arXiv preprint arXiv:1612.00410}, 2016.

\bibitem{he2016deep}
K.~He, X.~Zhang, S.~Ren, and J.~Sun, ``Deep residual learning for image recognition,'' in \emph{CVPR}, 2016, :770-8.

\bibitem{filiot2023scaling}
A.~Filiot \emph{et~al.}, ``Scaling self-supervised learning for histopathology with masked image modeling,'' \emph{medRxiv}, 2023, :2023-07.

\bibitem{goldman2020visualizing}
M.~J. Goldman \emph{et~al.}, ``Visualizing and interpreting cancer genomics data via the xena platform,'' \emph{Nat. Biotechnol.}, 2020, 38(6):675-8.

\bibitem{campanella2019clinical}
G.~Campanella \emph{et~al.}, ``Clinical-grade computational pathology using weakly supervised deep learning on whole slide images,'' \emph{Nat. Med.}, 2019, 25(8):1301-09.

\bibitem{otsu1975threshold}
N.~Otsu \emph{et~al.}, ``A threshold selection method from gray-level histograms,'' \emph{Automatica}, 1975, 11(285-296):23-7.

\bibitem{kingma2014adam}
D.~Kingma, ``Adam: a method for stochastic optimization,'' \emph{arXiv preprint arXiv:1412.6980}, 2014.

\bibitem{paszke2019pytorch}
A.~Paszke \emph{et~al.}, ``Pytorch: An imperative style, high-performance deep learning library,'' \emph{NeurIPS}, 2019, 32.

\bibitem{ilse2018attention}
M.~Ilse, J.~Tomczak, and M.~Welling, ``Attention-based deep multiple instance learning,'' in \emph{ICML}, 2018, :2127-36.

\bibitem{xiang2023exploring}
J.~Xiang and J.~Zhang, ``Exploring low-rank property in multiple instance learning for whole slide image classification,'' in \emph{ICLR}, 2023, pp. 333--351.

\bibitem{chen2022pan}
R.~J. Chen \emph{et~al.}, ``Pan-cancer integrative histology-genomic analysis via multimodal deep learning,'' \emph{Cancer Cell}, 2022, 40(8):865-78.

\bibitem{cerami2012cbio}
E.~Cerami \emph{et~al.}, ``The cbio cancer genomics portal: an open platform for exploring multidimensional cancer genomics data,'' \emph{Cancer discovery}, vol.~2, no.~5, pp. 401--404, 2012.

\bibitem{liberzon2015molecular}
A.~Liberzon, C.~Birger, H.~Thorvaldsd{\'o}ttir, M.~Ghandi, J.~P. Mesirov, and P.~Tamayo, ``The molecular signatures database hallmark gene set collection,'' \emph{Cell systems}, vol.~1, no.~6, pp. 417--425, 2015.

\bibitem{gillespie2022reactome}
M.~Gillespie \emph{et~al.}, ``The reactome pathway knowledgebase 2022,'' \emph{Nucleic acids research}, vol.~50, no.~D1, pp. D687--D692, 2022.

\bibitem{mcinnes2018umap}
L.~McInnes, J.~Healy, and J.~Melville, ``Umap: Uniform manifold approximation and projection for dimension reduction,'' \emph{arXiv preprint arXiv:1802.03426}, 2018.

\bibitem{sundararajan2017axiomatic}
M.~Sundararajan, A.~Taly, and Q.~Yan, ``Axiomatic attribution for deep networks,'' in \emph{International conference on machine learning}.\hskip 1em plus 0.5em minus 0.4em\relax PMLR, 2017, pp. 3319--3328.

\bibitem{narod2004brca1}
S.~A. Narod and W.~D. Foulkes, ``Brca1 and brca2: 1994 and beyond,'' \emph{Nature Reviews Cancer}, vol.~4, no.~9, pp. 665--676, 2004.

\bibitem{baselga2009novel}
J.~Baselga and S.~M. Swain, ``Novel anticancer targets: revisiting erbb2 and discovering erbb3,'' \emph{Nature Reviews Cancer}, vol.~9, no.~7, pp. 463--475, 2009.

\bibitem{paez2004egfr}
J.~G. Paez \emph{et~al.}, ``Egfr mutations in lung cancer: correlation with clinical response to gefitinib therapy,'' \emph{Science}, vol. 304, no. 5676, pp. 1497--1500, 2004.

\bibitem{dalgliesh2010systematic}
G.~L. Dalgliesh \emph{et~al.}, ``Systematic sequencing of renal carcinoma reveals inactivation of histone modifying genes,'' \emph{Nature}, vol. 463, no. 7279, pp. 360--363, 2010.

\bibitem{pena2012bap1}
S.~Pe{\~n}a-Llopis \emph{et~al.}, ``Bap1 loss defines a new class of renal cell carcinoma,'' \emph{Nature genetics}, vol.~44, no.~7, pp. 751--759, 2012.

\bibitem{kandoth2013mutational}
C.~Kandoth \emph{et~al.}, ``Mutational landscape and significance across 12 major cancer types,'' \emph{Nature}, vol. 502, no. 7471, pp. 333--339, 2013.

\bibitem{brien2012polycomb}
G.~L. Brien \emph{et~al.}, ``Polycomb phf19 binds h3k36me3 and recruits prc2 and demethylase no66 to embryonic stem cell genes during differentiation,'' \emph{Nature structural \& molecular biology}, vol.~19, no.~12, pp. 1273--1281, 2012.

\bibitem{asanuma2007protein}
K.~Asanuma \emph{et~al.}, ``Protein c inhibitor inhibits breast cancer cell growth, metastasis and angiogenesis independently of its protease inhibitory activity,'' \emph{International journal of cancer}, vol. 121, no.~5, pp. 955--965, 2007.

\bibitem{yang2025nfatc4}
W.~Yang \emph{et~al.}, ``Nfatc4 promotes lung adenocarcinoma progression via the ccnb1/cdk1 pathway and is a potential prognostic biomarker,'' \emph{Cancer Science}, 2025.

\bibitem{tang2023arpc1b}
Y.-F. Tang, B.~Qiao, Y.-B. Huang, and M.~Wang, ``Arpc1b is a novel prognostic biomarker for kidney renal clear cell carcinoma and correlates with immune infiltration,'' \emph{Frontiers in Molecular Biosciences}, vol.~10, p. 1202524, Sep. 2023.

\bibitem{alula2021nuclear}
K.~M. Alula, Y.~Delgado-Deida, D.~N. Jackson, K.~Venuprasad, and A.~L. Theiss, ``Nuclear partitioning of prohibitin 1 inhibits wnt/$\beta$-catenin-dependent intestinal tumorigenesis,'' \emph{Oncogene}, vol.~40, no.~2, pp. 369--383, 2021.

\bibitem{edwards2015cptac}
N.~J. Edwards \emph{et~al.}, ``The cptac data portal: a resource for cancer proteomics research,'' \emph{Journal of proteome research}, vol.~14, no.~6, pp. 2707--2713, 2015.

\bibitem{novak2018sensitivity}
R.~Novak, Y.~Bahri, D.~A. Abolafia, J.~Pennington, and J.~Sohl-Dickstein, ``Sensitivity and generalization in neural networks: an empirical study,'' \emph{arXiv preprint arXiv:1802.08760}, 2018.

\bibitem{dehkharghanian2023biased}
T.~Dehkharghanian \emph{et~al.}, ``Biased data, biased ai: deep networks predict the acquisition site of tcga images,'' \emph{Diagnostic pathology}, vol.~18, no.~1, p.~67, 2023.

\bibitem{weng2024grandqc}
Z.~Weng \emph{et~al.}, ``Grandqc: A comprehensive solution to quality control problem in digital pathology,'' \emph{Nature Communications}, vol.~15, no.~1, p. 10685, 2024.

\bibitem{janowczyk2019histoqc}
A.~Janowczyk, R.~Zuo, H.~Gilmore, M.~Feldman, and A.~Madabhushi, ``Histoqc: an open-source quality control tool for digital pathology slides,'' \emph{JCO clinical cancer informatics}, vol.~3, pp. 1--7, 2019.

\bibitem{kassab2024ffpe++}
M.~Kassab, M.~Jehanzaib, K.~Ba{\c{s}}ak, D.~Demir, G.~E. Keles, and M.~Turan, ``Ffpe++: Improving the quality of formalin-fixed paraffin-embedded tissue imaging via contrastive unpaired image-to-image translation,'' \emph{Medical Image Analysis}, vol.~91, p. 102992, Jan. 2024.

\bibitem{pinsker1960information}
M.~S. Pinsker, \emph{Information and Information Stability in Random Variables}.\hskip 1em plus 0.5em minus 0.4em\relax San Francisco, CA, USA: Holden-Day, 1960.

\bibitem{fano1961transmission}
R.~M. Fano, \emph{Transmission of Information: A Statistical Theory of Communications}.\hskip 1em plus 0.5em minus 0.4em\relax Cambridge, MA, USA: MIT Press, 1961, based on MIT lecture notes, 1950.

\bibitem{cover2006elements}
T.~M. Cover and J.~A. Thomas, \emph{Elements of Information Theory}, 2nd~ed.\hskip 1em plus 0.5em minus 0.4em\relax Hoboken, NJ, USA: Wiley-Interscience, 2006.

\end{thebibliography}

\end{document}